\documentclass[11pt]{article}

\usepackage[final]{acl}

\usepackage{times}
\usepackage{latexsym}

\usepackage[T1]{fontenc}

\usepackage[utf8]{inputenc}

\usepackage{microtype}

\usepackage{inconsolata}

\usepackage{graphicx}

\newcommand{\faHF}{\raisebox{-0.15em}{\includegraphics[height=1.2em]{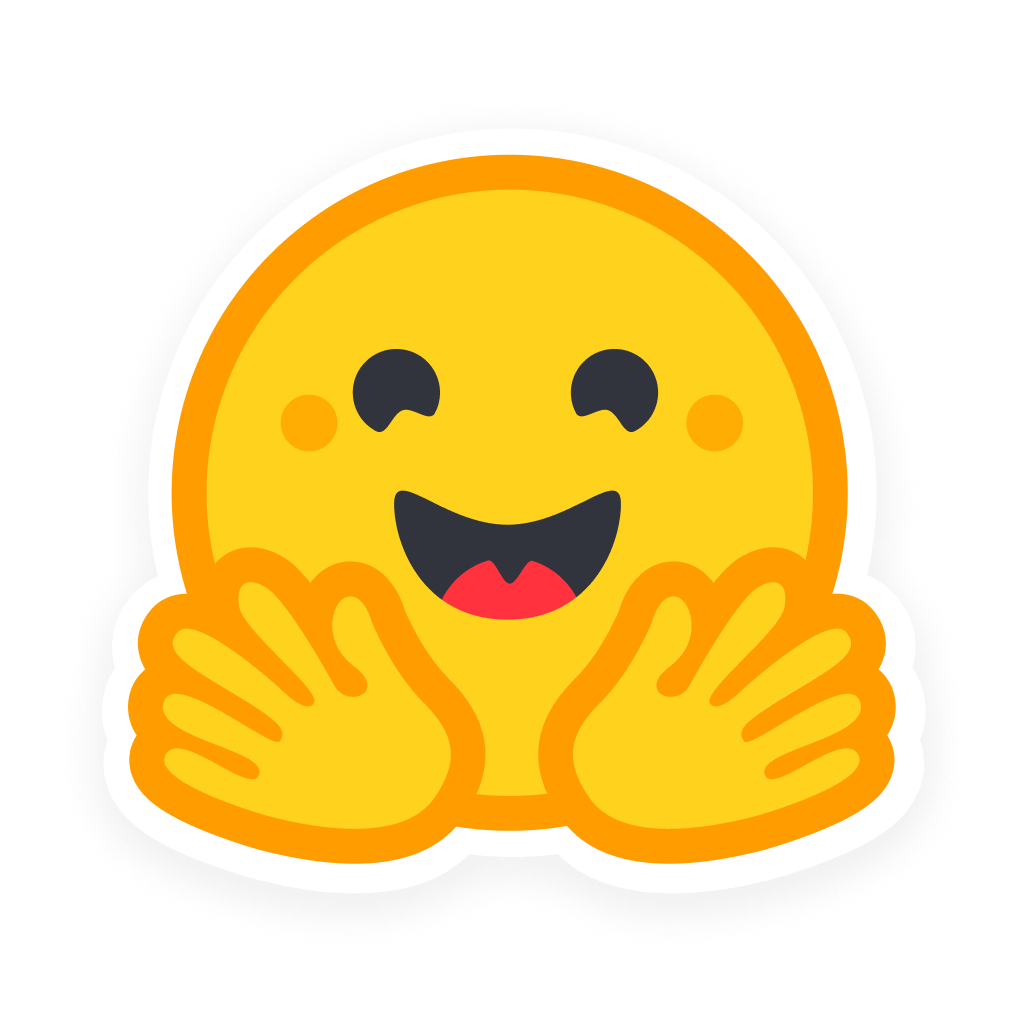}}}

\definecolor{dropgray}{gray}{0.45}

\title{ReactBench: A Benchmark for Topological Reasoning in MLLMs on Chemical Reaction Diagrams}

\author{
 \textbf{Qiang Xu\textsuperscript{1,2*}},
 \textbf{Shengyuan Bai\textsuperscript{1*}},
 \textbf{Yu Wang\textsuperscript{3}},
 \textbf{He Cao\textsuperscript{1}},
 \textbf{Leqing Chen\textsuperscript{1}},
\\
 \textbf{Yuanyuan Liu\textsuperscript{1}},
 \textbf{Bin Feng\textsuperscript{1}},
 \textbf{Zijing Liu\textsuperscript{1}},
 \textbf{Yu Li\textsuperscript{1$\dagger$}}
\\
 \textsuperscript{1}International Digital Economy Academy (IDEA)
\\
 \textsuperscript{2}Arizona State University
\\
 \textsuperscript{3}University of Wisconsin–Madison
\\
      \texttt{xuqiang@asu.edu}
      \quad
      \texttt{baishengyuan@idea.edu.cn}
\\
\faHF \href{https://huggingface.co/datasets/IDEA-AI4S/ReactBench}{\texttt{ReactBench}}
}

\usepackage{url}
\usepackage{graphicx}
\usepackage{placeins}
\usepackage{booktabs}
\usepackage{adjustbox}
\usepackage{multirow}
\usepackage{colortbl}
\usepackage{wrapfig}
\usepackage{caption}
\usepackage{xspace}
\usepackage{float}
\usepackage{makecell}
\usepackage{ragged2e}
\usepackage{tabularx}
\usepackage{array}
\newcommand{\ours}{ReactBench\xspace}
\newcommand{\oursdata}{ChemReaction\xspace}

\usepackage{listings}
\usepackage{tcolorbox}
\tcbuselibrary{listings,breakable}

\lstdefinelanguage{json}{
  basicstyle=\ttfamily\small,
  showstringspaces=false,
  breaklines=true,
  morestring=[b]",
  morecomment=[l]{//},
  morecomment=[s]{/*}{*/},
  sensitive=true,
  literate=
   *{true}{{{\color{blue}true}}}{4}
    {false}{{{\color{blue}false}}}{5}
    {null}{{{\color{blue}null}}}{4}
}

\newtcblisting{json}{
  listing engine=listings,
  listing only,
  breakable,
  colback=gray!3,
  colframe=gray!50,
  boxrule=2pt,
  arc=0.5pt,
  fonttitle=\small,
  title=\textbf{Example output from RxnScribe},
  listing options={
    language=json,
    basicstyle=\ttfamily\small,
    columns=fullflexible,
    numbers=left,
    numberstyle=\tiny,
    numbersep=5pt,
    showstringspaces=false,
    breaklines=true
  }
}

\newtcblisting{gt}{
  listing engine=listings,
  listing only,
  breakable,
  colback=gray!3,
  colframe=gray!50,
  boxrule=2pt,
  arc=0.5pt,
  fonttitle=\small,
  title=\textbf{Example of Ground-Truth JSON},
  listing options={
    language=json,
    basicstyle=\ttfamily\small,
    columns=fullflexible,
    numbers=left,
    numberstyle=\tiny,
    numbersep=5pt,
    showstringspaces=false,
    breaklines=true
  }
}

\begin{document}
\twocolumn[{
	\renewcommand\twocolumn[1][]{#1}
	\maketitle
    \vspace{-10pt}
	\begin{center} 
		\includegraphics[width=\linewidth]{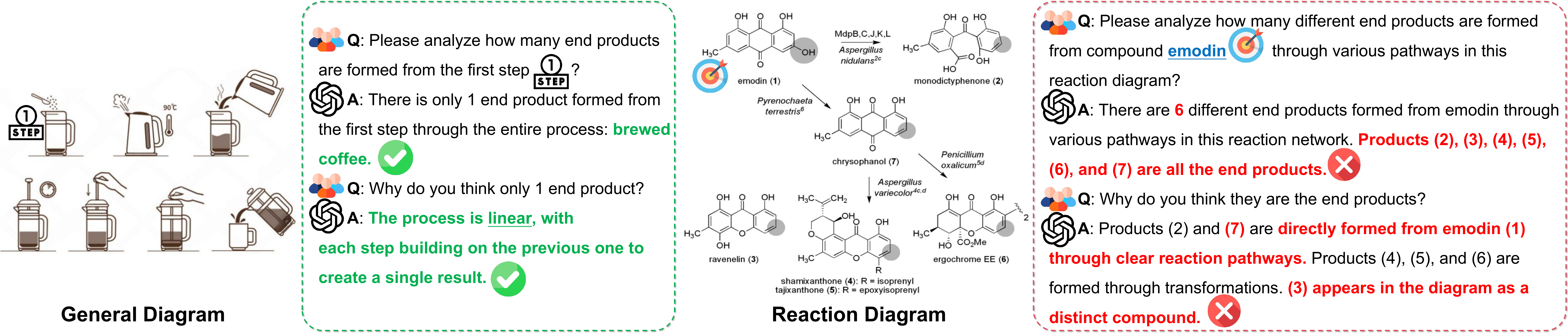}
		\captionsetup{type=figure} 
        \vspace{-20pt}
        \caption{\textbf{Comparing General Diagram and Chemical Reaction Diagram on Element Localization Tasks.}
  GPT-4o performs well on general diagrams for identification tasks. However, for chemical reaction diagram, the structural complexity of molecular diagrams and reaction pathways leads to errors.}
       \label{fig:motivation}
	\end{center}
}]

\begingroup
\renewcommand{\thefootnote}{}
\footnotetext{
\textsuperscript{*} Equal contribution.
\textsuperscript{$\dagger$} Corresponding author.
}
\addtocounter{footnote}{-1}
\endgroup

\begin{abstract}
Multimodal Large Language Models (MLLMs) excel at recognizing individual visual elements and reasoning over simple linear diagrams. However, when faced with complex topological structures involving branching paths, converging flows, and cyclic dependencies, their reasoning capabilities degrade sharply, even on tasks as basic as counting endpoints. Existing benchmarks fail to probe this gap, focusing on semantic comprehension rather than structural reasoning.
We introduce \textbf{ReactBench}, a benchmark that reveals fundamental limitations in structural reasoning through chemical reaction diagrams. These real-world scientific diagrams offer an ideal testbed because they naturally span diverse structures from linear chains to cyclic graphs, while requiring both precise local recognition and coherent global reasoning. Our benchmark comprises 1,618 expert-annotated QA pairs across four hierarchical task dimensions.
Extensive evaluation across 24 MLLMs reveals a significant performance gap exceeding 30\% between anchor-based tasks and holistic structural reasoning tasks.
Controlled ablations confirm this bottleneck lies in reasoning, not perception. These findings expose a fundamental deficit in structural understanding and establish directions for advancing visual reasoning. 
\end{abstract}

\section{Introduction}
\label{sec:intro}
Recent advances in MLLMs have demonstrated remarkable capabilities in visual-linguistic reasoning across diverse domains~\cite{gpt4o,claude35,gemini15,qwen25vl}. These models excel at recognizing individual visual elements and reasoning over simple linear diagrams. However, real-world diagrams often encode information through complex topological structures involving branching paths, converging flows, and cyclic dependencies. The capacity of MLLMs to reason about such structural connections remains largely unexplored, as existing benchmarks~\cite{docvqa,textvqa,chartqa,scienceqa} primarily evaluate semantic comprehension rather than structural reasoning capabilities.

\begin{figure*}[!t]
    \centering    
    \includegraphics[width=\linewidth]{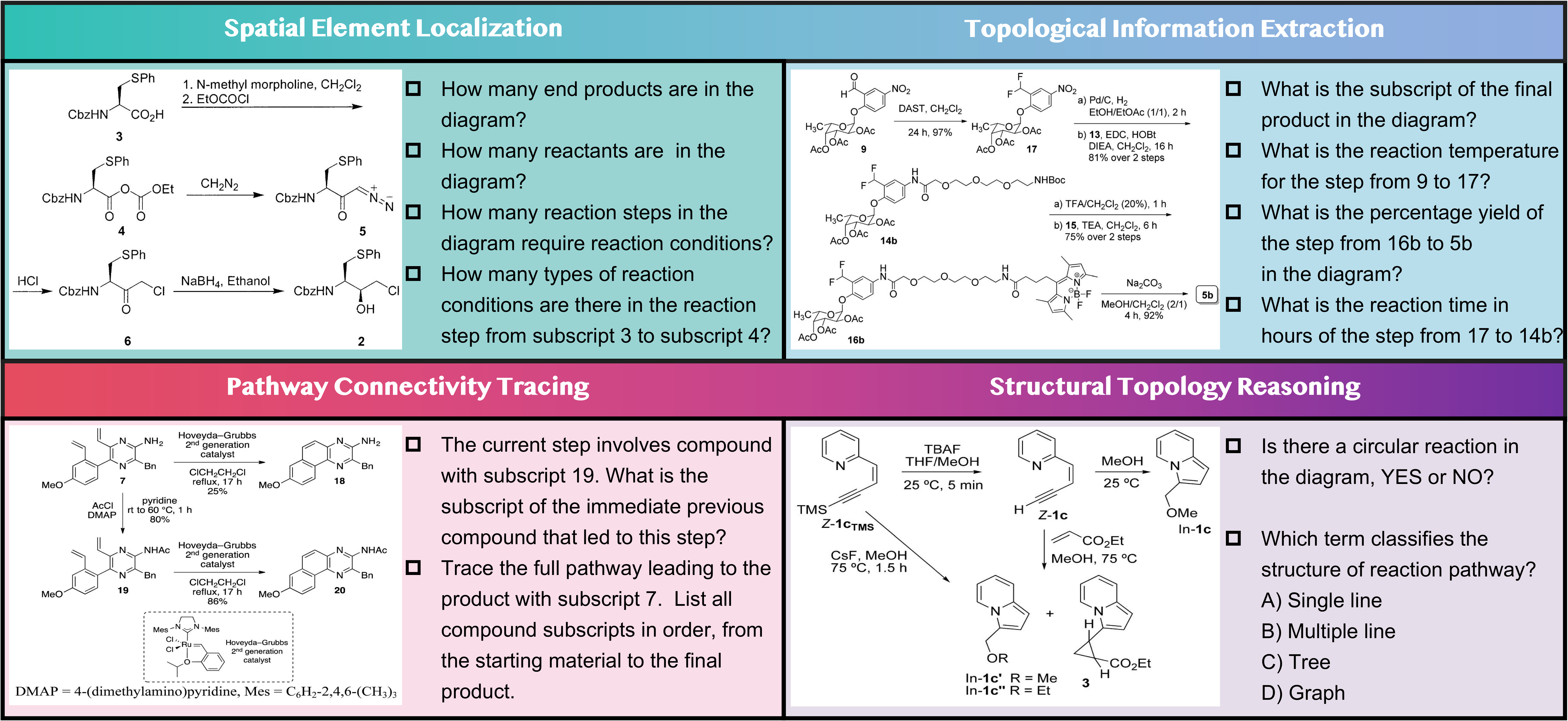}
    \vspace{-15pt}
    \caption{\textbf{Overview of \ours.} \ours systematically evaluates topological reasoning on chemical reaction diagrams across four complexity-stratified dimensions.}
    \label{fig:overview}
    \vspace{-14pt}
\end{figure*}

We illustrate this gap through a controlled comparison in Fig.~\ref{fig:motivation}. Both tasks require the same topological operation of identifying terminal nodes in a diagram. When presented with a simple linear topology such as a sequential coffee brewing process, GPT-4o correctly identifies the single endpoint. However, when the same task is posed on a branching chemical network, the model fails systematically, misidentifying six endpoints instead of the actual three. Notably, neither task requires domain expertise. The failure is attributable to an inability to reason about complex branching structures rather than a lack of chemical knowledge.

These observations reveal a fundamental limitation in current MLLMs. The ability to perceive local elements does not guarantee global structural understanding. As topological complexity increases, models systematically fail to integrate visual elements into coherent global structures.

Diagnosing this limitation requires a testbed that combines diverse topological structures with demands for both local recognition and global reasoning. 
Chemical reaction diagrams provide such a testbed: they span structures from linear chains to cyclic graphs, while requiring accurate perception of molecules, conditions, and reaction connectivity.

We introduce \textbf{\ours}, a benchmark designed to diagnose structural reasoning limitations in MLLMs through chemical reaction diagrams. \ours comprises 1,618 question-answer pairs carefully curated from real-world chemistry literature and patents, organized into four hierarchical task dimensions: spatial element localization, topological information extraction, pathway connectivity tracing, and structural topology reasoning. This design enables precise identification of where model capabilities degrade along the perception-to-reasoning spectrum. Our main contributions are: 
\begin{itemize}

\item We present ReactBench, the first benchmark that systematically diagnoses structural reasoning limitations in MLLMs through hierarchical task decomposition, comprising 1,618 expertly annotated question-answer pairs across four skill dimensions.
\item We identify a persistent gap between local diagram understanding and global structural reasoning in current MLLMs. Although models perform well on anchor-based perception and local pathway tracing, they struggle to integrate local visual cues into coherent topological representations, showing that structural complexity remains a key challenge.
\item Through controlled ablations across multiple model families, we show that the main bottleneck lies in reasoning rather than perception, and identify recurring structural failure modes that inform future visual reasoning systems.
\end{itemize}

\section{Related Work}
\label{sec:related}
\subsection{Multimodal Large Language Models}
Multimodal Large Language Models (MLLMs) extend text-only LLMs by integrating visual comprehension capabilities. Early works established fundamental vision-language alignment (e.g., CLIP~\cite{clip}) and cross-modal fusion (e.g., BLIP-2~\cite{li2023blip2}). Through large-scale pre-training and instruction tuning on diverse multimodal datasets, the field has rapidly advanced, achieving remarkable proficiency in general visual reasoning and Optical Character Recognition (OCR) understanding. Today, the landscape is driven by highly capable open-source models (e.g., Qwen3-VL~\cite{qwen3vl}, Google Gemma 4~\cite{gemma4}, and Kimi-K2.6~\cite{kimi-k2.6}), alongside powerful proprietary models (e.g., GPT-5.5~\cite{gpt55}, Claude-Sonnet-4.6~\cite{claudesonnet46}, and Gemini-3.1-Pro~\cite{gemini31pro}). However, while these state-of-the-art MLLMs excel at recognizing individual visual elements and extracting localized information, their capacity to perform holistic structural reasoning~\cite{chemfp, chemo,mozi} over complex topological diagrams remains largely unexplored.


\begin{figure}[!t]
    \centering
    {
    \resizebox{0.9\linewidth}{!}{
    \begin{tabular}{l r}
        \toprule
        \textbf{Category} & \textbf{Number} \\
        \midrule
        \multicolumn{2}{l}{\textbf{Question Dimensions}} \\
        Spatial Element Localization & 835 \\
        Topological Information Extraction & 414 \\
        Pathway Connectivity Tracing &  167 \\
        Structural Topology Reasoning &  202 \\
        \midrule
        \multicolumn{2}{l}{\textbf{Question types}} \\
        Numerical questions & 1094 \\
        Multiple-choice questions & 202 \\
        Free-form questions & 322 \\
        \midrule
        \textbf{Total} & \textbf{1618} \\
        \bottomrule
    \end{tabular}
    }
    \captionof{table}{Key statistics of \oursdata.}
    \label{tab:keystats}
    }

    \vspace{6pt} 

    \includegraphics[width=0.9\linewidth]{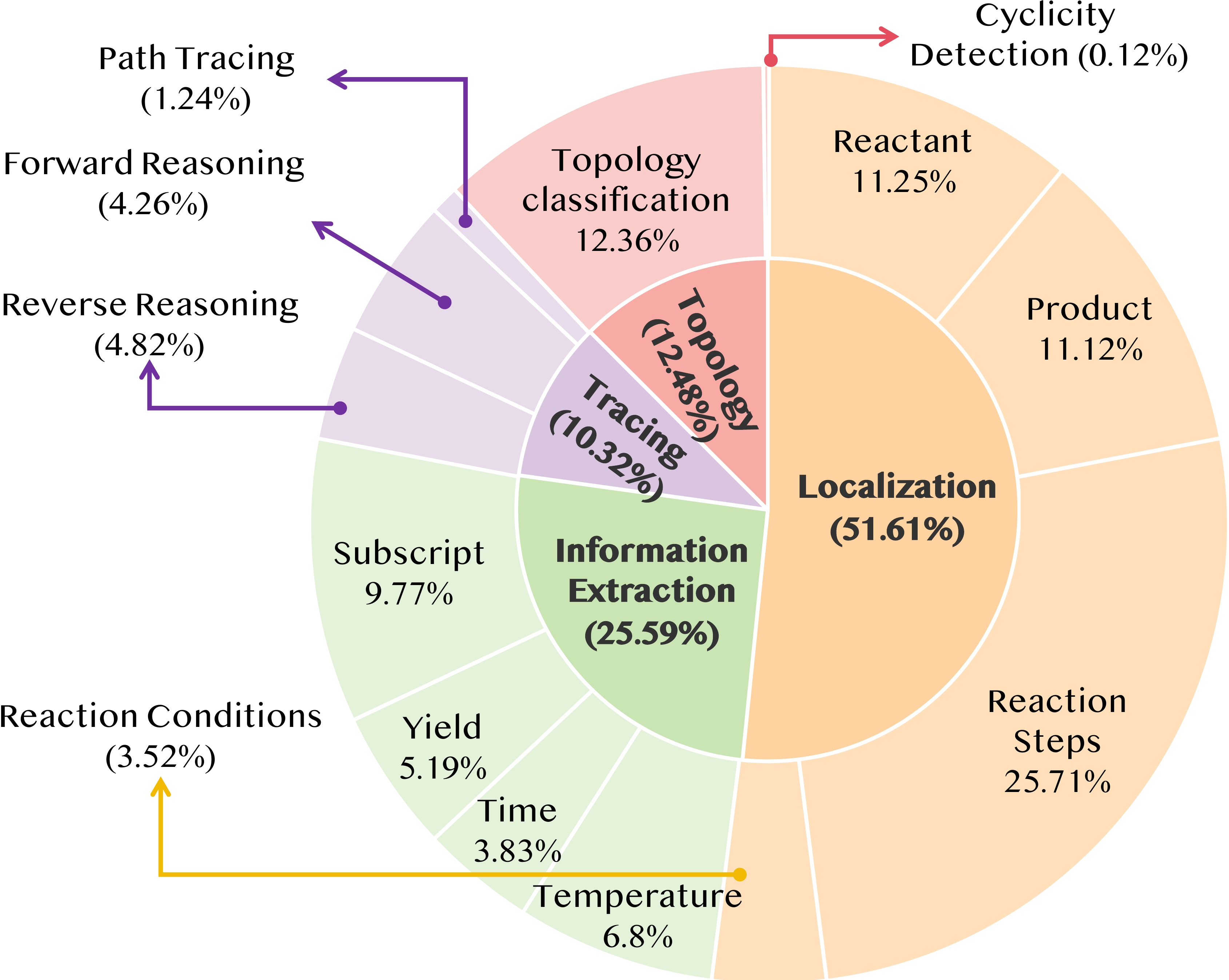}
    \caption{Composition of \oursdata.}
    \label{fig:stat}
\vspace{-16pt}
\end{figure}

\subsection{Visual Question Answering Benchmarks}
\label{subsec:vqa_benchmarks}

Visual Question Answering (VQA) benchmarks have been instrumental in driving the development and evaluation of Multimodal Large Language Models. Existing datasets like DocVQA~\cite{docvqa} and ChartQA~\cite{chartqa} primarily evaluate visual information extraction capabilities across relatively simple, standardized layouts~\cite{li2025enhancing,counterfactual}. On the other hand, science-oriented benchmarks such as ScienceQA~\cite{scienceqa} and AI2D~\cite{ai2d} test domain-specific scientific knowledge and basic diagrammatic understanding. 
Chemical-domain benchmarks such as MoleculeQA~\cite{moleculeqa} and ChemCoTBench~\cite{beyond} further evaluate molecular comprehension and chemical reasoning, respectively. 
However, these benchmarks do not systematically evaluate structural reasoning over complex topological diagrams. In these scenarios, visual inputs typically act as static semantic references rather than intricate relational networks~\cite{bai2025enhancing}. Consequently, while current VQA benchmarks heavily focus on general semantic comprehension and basic grounding, they lack a systematic evaluation of complex topological reasoning. Our work addresses this critical gap by specifically isolating and evaluating multi-hop, graph-theoretic reasoning within dense structural topologies.

\begin{figure*}[t]
    \centering    
    \includegraphics[width=0.9\linewidth]{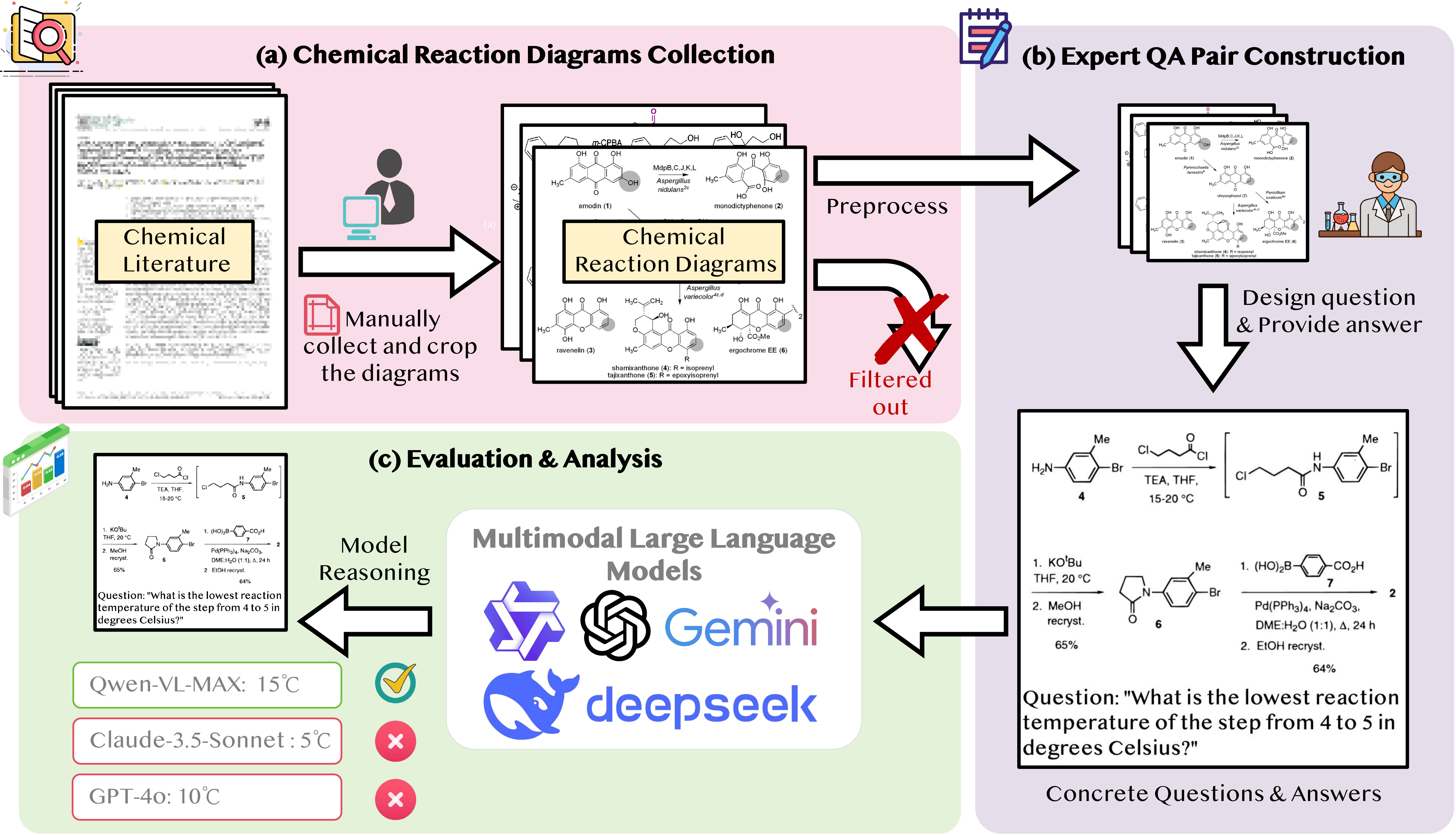}
    \vspace{-6pt}
    \caption{\textbf{ReactBench pipeline.} We collect and curate chemical reaction diagrams from literature and patents, construct expert-validated QA pairs, and evaluate MLLMs across predefined topological reasoning tasks.}
    \label{fig:pipeline}
    \vspace{-10pt}
\end{figure*}

\subsection{Chemical Reaction Diagram Understanding}

Existing approaches to chemical reaction diagram understanding employ multi-stage pipelines that first perform Optical Chemical Structure Recognition (OCSR) to detect individual molecular structures, then convert detected elements into text-based representations such as the Simplified Molecular Input Line Entry System (SMILES)~\cite{decimer,molscribe,rxnscribe,img2mol}. Representative methods include RxnScribe~\cite{rxnscribe} and ReactionDataExtractor 2.0~\cite{wilary2023reactiondataextractor}. Such pipeline-based frameworks are fundamentally incompatible with end-to-end MLLMs, which directly process visual inputs without intermediate symbolic conversion. Furthermore, text-based representations discard spatial positioning and pathway directionality, precluding differentiation between recognition errors and reasoning deficits~\cite{optical,chemgrapher,decimer_ai}. In contrast, ReactBench evaluates MLLMs' topological reasoning capabilities through visual question-answering tasks, enabling direct assessment of structural understanding without pipeline-induced confounds. \looseness=-1

\section{The \ours Benchmark}
\label{sec:benchmark}

We introduce \underline{\ours}, a specialized benchmark designed to evaluate MLLMs' ability to comprehend and reason about chemical reaction diagrams. \ours consists of a high-quality dataset \underline{\oursdata} with 1,618 rigorously curated QA pairs derived from real-world chemical reaction diagrams, and a comprehensive evaluation framework assessing four key dimensions of reaction diagram understanding: \textbf{(1) spatial element localization}, \textbf{(2) topological information extraction}, \textbf{(3) pathway connectivity tracing}, \textbf{(4) structural topology reasoning}. Key statistics of \oursdata are summarized in Tab.~\ref{tab:keystats}, and its composition is illustrated in Fig.~\ref{fig:stat}. The complete benchmark construction pipeline is depicted in Fig.~\ref{fig:pipeline}. Below, we detail the dataset collection and annotation methodology for \oursdata, and the evaluation scope of \ours.

\subsection{Dataset Details}

\noindent\textbf{Data Collection and Curation.}
\ours is constructed from over 1,300 unique real-world chemical reaction diagrams curated from chemistry literature and patent databases, ensuring both academic rigor and practical relevance. 
The academic sources comprise diagrams extracted from articles published between 1996 and 2016 in major Chemical Society journals, including \textit{The Journal of Organic Chemistry}, \textit{Journal of the American Chemical Society}, and \textit{Organic Letters}. 
To ensure diversity, each diagram is sourced from a distinct article, covering varied layouts, notation styles, and information densities. 
We further filter out diagrams with incomplete mechanisms, ambiguous visual structures, or duplicate content. 
The final dataset contains 1,618 question-answer pairs.

\noindent\textbf{Taxonomy of Reaction Topologies.}
To capture diverse visual complexity and reasoning challenges, we stratify the curated diagrams into four topology categories:
(\textbf{a}) \textit{Single-line}: linear pathways with unidirectional flows;
(\textbf{b}) \textit{Multiple-lines}: parallel or branched mechanisms with competing intermediates;
(\textbf{c}) \textit{Tree}: hierarchical branching with divergent synthesis routes;
(\textbf{d}) \textit{Graph}: cyclic networks requiring non-linear reasoning.

\subsection{Question Design}
To evaluate MLLMs' chemical reaction diagram understanding, we design questions across four dimensions, as shown in Fig.~\ref{fig:overview}.

\noindent\textbf{Spatial Element Localization.} 
This task evaluates fine-grained spatial reasoning within structured diagrams. 
Models must identify node roles, such as sources, sinks, and intermediates, based on topological position and visual context.

\noindent\textbf{Topological Information Extraction.} 
This dimension assesses structured information extraction under topological context. 
Models must extract numerical and textual attributes, such as reaction conditions, yields, and subscripts, while preserving their correspondence to graph elements.

\noindent\textbf{Pathway Connectivity Tracing.} 
This dimension probes connectivity and path traversal in reaction networks. 
Models must trace pathways, predict sequential nodes, and reconstruct complete paths from topological connections alone, testing graph-theoretic reasoning rather than chemical prediction.

\noindent\textbf{Structural Topology Reasoning.} 
This dimension evaluates holistic understanding of reaction topology. 
Models must classify global structural patterns, such as linear, branched, cyclic, tree-like, or graph topologies, and detect properties such as cycles or terminal nodes.

\begin{table*}[!t]
\caption{Detailed evaluation results on \ours across different models, showing the answer accuracy (Acc.) of each model in each task. \textbf{Average} denotes the arithmetic mean of a model’s scores over all four tasks. The \textbf{best} and \underline{second-best} per model category is highlighted with bold and underlined respectively.}
\label{tab:performance}
\vspace{-5pt}
\centering
\renewcommand{\arraystretch}{1.15} 
\setlength{\tabcolsep}{4.5pt} 

\begin{adjustbox}{max width=\textwidth}
\begin{tabular}{ccccccc} 
\toprule
\multirow{2}{*}{\textbf{Models}} & \multirow{2}{*}{\textbf{\# Params}} & \multicolumn{4}{c}{\textbf{\ours}} & \multirow{2}{*}{\textbf{Average}} \\ 
\cline{3-6}
                                 &                                  & \textbf{Localization} & \textbf{Extraction} & \textbf{Tracing} & \textbf{Reasoning} & \\ 
\hline
\multicolumn{7}{l}{{\cellcolor[rgb]{0.976,0.976,0.976}}\textit{Proprietary Models}} \\ 
\hline
GPT-5.5                           & -                                & 55.21                    & \underline{88.41}               & 83.23              & 51.49 & \underline{69.59} \\
GPT-4o                           & -                                & 44.07                    & 79.71               & 82.04              & 49.50 & 63.83 \\
Claude-Sonnet-4.6                & -                                & \underline{55.69}           & 75.36   & \underline{90.42}     & 46.53 & 67.00 \\
Claude-Sonnet-3.5                & -                                & 50.78           & 81.88   & \textbf{92.22}     & 49.50 & 68.60 \\
Gemini-3.1-Pro                   & -                                & \textbf{59.16}                    & 85.27               & 86.23              & \textbf{64.85}    & \textbf{73.88} \\
Gemini-1.5-Pro                   & -                                & 45.75                    & 75.60               & 71.86              & \underline{55.45}    & 62.17 \\
Qwen-VL-MAX                      & -                                & 49.22        & \textbf{90.10}      & 88.62  & 32.18             & 65.03 \\ 
\hline
\multicolumn{7}{l}{{\cellcolor[rgb]{0.976,0.976,0.976}}\textit{Open Source Models}} \\ 
\hline
InternVL2.5-MPO                  & 8B                               & 38.20                    & 66.43               & 59.28              & 37.62             & 50.38 \\
InternVL2.5                      & 8B                               & 27.43                    & 65.22               & 45.51              & 26.73             & 41.22 \\
InternVL2.5                      & 26B                              & 33.89                    & 66.18               & 65.87              & 19.80             & 46.44 \\
InternVL2.5                      & 78B                              & 41.08                    & 79.47               & 67.66  & 26.73             & 53.74 \\ 
\hline
Qwen3-VL-Instruct                       & 8B                               & 52.34                    & 85.02   & 82.04              & 32.67             & 63.02 \\
Qwen3-VL-Instruct                       & 32B                               & \underline{60.24}                    & 86.96   & \textbf{91.02}              & \textbf{57.43}             & \textbf{73.91} \\
Qwen3-VL-Thinking                       & 8B                               & 49.22                    & 85.99   & 76.65              & 29.11             & 60.24 \\
Qwen3-VL-Thinking                       & 32B                               & \textbf{63.59}                    & 88.89   & \underline{88.02}              & 43.07             & \underline{70.89} \\
\hline
Qwen2.5-VL                       & 3B                               & 24.43                    & \underline{89.61}   & 49.70              & 34.65             & 49.60 \\
Qwen2.5-VL                       & 7B                               & 37.96                    & 85.75               & 66.47              & 39.11             & 57.32 \\
Qwen2.5-VL                       & 72B                              & 46.47           & \textbf{89.86}      & 83.83     & \underline{54.46}    & 68.66 \\ 
\hline
DeepSeek-VL2                     & 3B                               & 45.39        & 50.72               & 16.17              & 33.66             & 36.49 \\
DeepSeek-VL2                     & 16B                              & 28.62                    & 78.50               & 8.38               & 28.71             & 36.05 \\ 
\hline
LLaVA-NeXT-Mistral               & 7B                               & 18.44                    & 30.43               & 26.95              & 21.78             & 24.40 \\
LLaVA-NeXT-Vicuna                & 13B                              & 13.77                    & 39.13               & 1.80               & 28.22             & 20.73 \\ 
\hline
Phi-3.5-vision-instruct          & 4.2B                             & 21.80                    & 69.75               & 29.34              & 38.12             & 39.75 \\
MiniCPM-o 2.6                    & 8B                               & 35.09                    & 75.12               & 51.50              & 39.60 & 50.33 \\
\bottomrule
\end{tabular}
\end{adjustbox}
\vspace{-15pt}
\end{table*}

\subsection{Evaluation Method}
To simplify the evaluation process, all questions in \oursdata are presented in either multiple-choice or open-ended formats with concise, easily verifiable ground truth answers. We utilize prompting and template matching to extract answers. Prompts guide the model in generating responses in both full and short-answer formats. After generations, the short answer is extracted to compare it with the ground truth. Detailed prompts used in our experiments can be found in Appendix~\ref{app:prompt}. This design eliminates the need for using LLM-as-a-judge~\cite{chen2024mllm} and enables fully automated accuracy measurement through exact string matching.

\section{Experiment}
\label{sec:experiment}

\vspace{-5pt}
\subsection{Experimental Setups}
We comprehensively evaluate various MLLMs across multiple chemical reaction diagram understanding tasks. We test open-source models including Qwen3-VL-Instruct (8B, 32B) and Qwen3-VL-Thinking (8B, 32B)~\cite{qwen3vl}, Qwen2.5-VL (3B, 7B, 72B)~\cite{qwen25vl}, InternVL2.5 (8B-MPO, 8B, 26B, 78B)~\cite{internvl25}, LLaVA-NeXT (Mistral-7B, Vicuna-13B)~\cite{liu2024llavanext}, MiniCPM-o 2.6 (8B)~\cite{minicpmo}, DeepSeek-VL2 (3B, 16B)~\cite{deepseekvl2} and Phi-3.5-vision-instruct (4.2B)~\cite{phi35v}. Furthermore, we evaluate proprietary models such as GPT-5.5~\cite{gpt55}, GPT-4o~\cite{gpt4o}, Claude-Sonnet-4.6~\cite{claudesonnet46}, Claude-Sonnet-3.5~\cite{claude35}, Gemini-3.1-Pro~\cite{gemini31pro}, Gemini-1.5-Pro~\cite{gemini15}, and Qwen-VL-MAX~\cite{qwenvlmax}. Evaluation of these MLLMs ensures fair comparison considering model architecture differences and parameter scales.

\begin{figure*}[!t]
    \centering    
    \includegraphics[width=\linewidth]{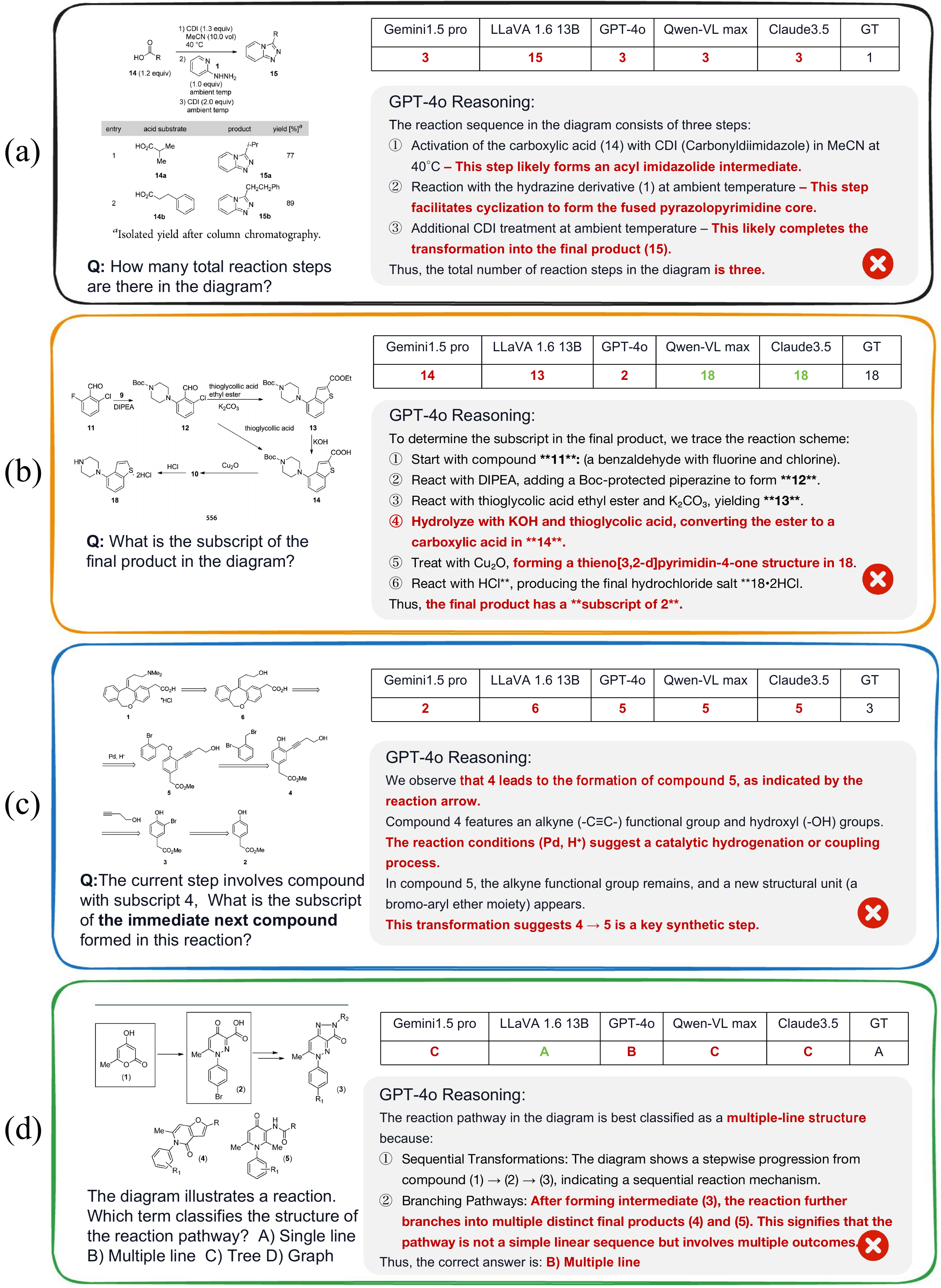}
    \vspace{-15pt}
    \caption{Qualitative analysis of recurring failure patterns in ReactBench evaluation. Each case study exemplifies a characteristic reasoning deficit observed across multiple model architectures.}
    \label{fig:casestudy}
    \vspace{-10pt}
\end{figure*} 

\subsection{Main Results}
In this section, we evaluate both open-source and proprietary MLLMs on \ours. 
Tab.~\ref{tab:performance} reports model accuracy across four tasks. 
The results reveal a clear pattern: recent frontier MLLMs substantially improve the overall performance ceiling, yet a clear gap remains between anchor-based perception and holistic topological reasoning.

\noindent\textbf{Recent frontier models improve the performance ceiling, but do not close the topology gap.}
The strongest models in our evaluation substantially outperform earlier MLLMs on \ours. 
Among all evaluated models, Qwen3-VL-Instruct-32B achieves the highest average accuracy of 73.91\%, closely followed by Gemini-3.1-Pro with 73.88\%. 
Qwen3-VL-Thinking-32B and GPT-5.5 also obtain strong average scores of 70.89\% and 69.59\%, respectively. 
These results indicate that recent MLLMs have made notable progress in chemical reaction diagram understanding. 
However, their gains are uneven across task types. 
The best scores on Extraction and Tracing reach 90.10\% and 92.22\%, respectively, while the best scores on Localization and Reasoning remain lower at 63.59\% and 64.85\%. 
This gap suggests that local visual recognition and pathway following are more mature than global topological abstraction.

\noindent\textbf{Thinking does not consistently improve topological reasoning.}
A notable finding is that thinking-style models do not uniformly outperform their non-thinking counterparts. 
For proprietary models, GPT-5.5 improves over GPT-4o in average accuracy, increasing from 63.83\% to 69.59\%. 
Gemini-3.1-Pro also substantially improves over Gemini-1.5-Pro, increasing from 62.17\% to 73.88\%, and achieves the best Reasoning accuracy among all models at 64.85\%. 
However, this trend is not universal. 
Claude-Sonnet-4.6 obtains a lower average score than Claude-Sonnet-3.5, and also slightly underperforms it on both Tracing and Reasoning.

The comparison between Qwen3-VL-Instruct and Qwen3-VL-Thinking provides a more controlled view of this effect. 
Although Qwen3-VL-Thinking-32B achieves higher Localization and Extraction scores than Qwen3-VL-Instruct-32B, it performs substantially worse on Reasoning, dropping from 57.43\% to 43.07\%. 
A similar pattern appears at the 8B scale, where Qwen3-VL-Thinking-8B underperforms Qwen3-VL-Instruct-8B in average accuracy. 
These results suggest that explicit thinking-style inference may improve localized inspection or element-level grounding, but does not necessarily lead to better global topology abstraction. 
In some cases, the model may still rely on local visual evidence without successfully integrating it into a coherent global reaction structure.

\noindent\textbf{Local pathway tracing does not imply global structural reasoning.}
Another important pattern is the large gap between Tracing and Reasoning within the same model. 
For example, Claude-Sonnet-4.6 achieves 90.42\% on Tracing but only 46.53\% on Reasoning. 
Qwen-VL-MAX reaches 88.62\% on Tracing but drops to 32.18\% on Reasoning. 
Similarly, Qwen3-VL-Thinking-32B obtains 88.02\% on Tracing but only 43.07\% on Reasoning. 
These results indicate that following local connectivity between adjacent nodes and inferring the global topology of the entire reaction network require different capabilities. 
A model may correctly trace local reaction paths while still failing to classify the overall reaction topology.

\noindent\textbf{Open-source frontier models are competitive with proprietary models.}
The updated evaluation also shows that recent open-source models can match or even surpass proprietary systems on \ours. 
Qwen3-VL-Instruct-32B achieves the best overall average accuracy among all evaluated models, slightly outperforming Gemini-3.1-Pro. 
It also obtains 91.02\% on Tracing and 57.43\% on Reasoning, outperforming most proprietary models on these dimensions. 
This suggests that the performance gap between open-source and proprietary MLLMs has become much smaller for chemical reaction diagram understanding, particularly on tasks involving local pathway following and moderately structured visual reasoning.

\begin{table}[!t]
\caption{Performance comparison of Qwen2.5-VL using reaction-diagram images and linearized ground-truth JSON. \textbf{Loc.}: localization; \textbf{Ext.}: extraction; \textbf{Trac.}: tracing; \textbf{Reas.}: reasoning.}
\vspace{-6pt}
\centering
\small
\setlength{\tabcolsep}{3.3pt}
\begin{tabular}{llcccc}
\toprule
\textbf{Model} & \textbf{Input} & \textbf{Loc.} & \textbf{Ext.} & \textbf{Trac.} & \textbf{Reas.} \\
\midrule
\multirow{2}{*}{Qwen2.5-VL-3B} 
& Image     & 24.43          & \textbf{89.61} & \textbf{49.70} & \textbf{34.65} \\
& Text-Only & \textbf{34.37} & 34.54          & 19.17          & 9.90 \\
\midrule
\multirow{2}{*}{Qwen2.5-VL-7B} 
& Image     & 37.96          & \textbf{85.75} & \textbf{66.47} & \textbf{39.11} \\
& Text-Only & \textbf{39.88} & 41.06          & 10.78          & 8.91 \\
\bottomrule
\end{tabular}
\label{tab:text_vs_image}
\vspace{-14pt}
\end{table}

\noindent\textbf{Extraction is approaching saturation, while topology remains the main bottleneck.}
Topological information extraction shows high performance across many strong models. 
Qwen-VL-MAX achieves 90.10\%, Qwen2.5-VL-72B achieves 89.86\%, Qwen2.5-VL-3B achieves 89.61\%, and Qwen3-VL-Thinking-32B achieves 88.89\%. 
This indicates that local anchor recognition, including textual and numerical information extraction, is no longer the primary bottleneck for frontier MLLMs. 
By contrast, holistic tasks such as endpoint localization and structural topology reasoning remain substantially more challenging. 
Therefore, the key limitation of current MLLMs lies not merely in recognizing individual diagram elements, but in integrating these elements into a coherent global representation of the reaction topology.

\section{Diagnosing the Reasoning Gap: Analysis and Ablations}
\label{sec:analysis_and_ablations}

The preceding results show that strong models can exceed 80\% on extraction and tracing, yet still struggle with endpoint localization and topology reasoning. This gap raises a key question: \textit{Does the bottleneck stem from visual perception or structural reasoning?} We investigate this through targeted diagnostic experiments.

\subsection{Bottleneck: Perception or Reasoning}

The preceding results reveal a clear performance gap. We now investigate whether this deficit originates from the visual modality itself, a failure in perception, or a limitation in reasoning. 

\noindent\textbf{Step 1: The Necessity of Visual Modality.} We first investigate whether complex topological reasoning can be bypassed by providing perfectly structured text. We conduct an ablation study using ground-truth structured JSON (see Appendix~\ref{app:gt_json}), linearized into a text-only prompt to completely replace the image input. This JSON representation encompasses all essential chemical information, including SMILES strings, molecular names, textual tags (e.g., A, B, C), bounding box coordinates (bbox), reaction roles, and reaction conditions. As shown in Tab.~\ref{tab:text_vs_image}, while Localization scores slightly improve due to the explicit coordinates in the JSON, performance on Extraction, Tracing, and Reasoning tasks drops precipitously. Notably, for Qwen2.5-VL-7B, the Tracing score falls from 66.47\% to a mere 10.78\%. This contrast demonstrates that topological reasoning inherently requires the spatial and structural information encoded in visual diagrams, which is lost during the linearization process. However, this necessity of visual input does not imply the bottleneck lies in perception, a point we investigate next.

\begin{figure}[!t]
    \centering
    \includegraphics[width=0.8\linewidth]{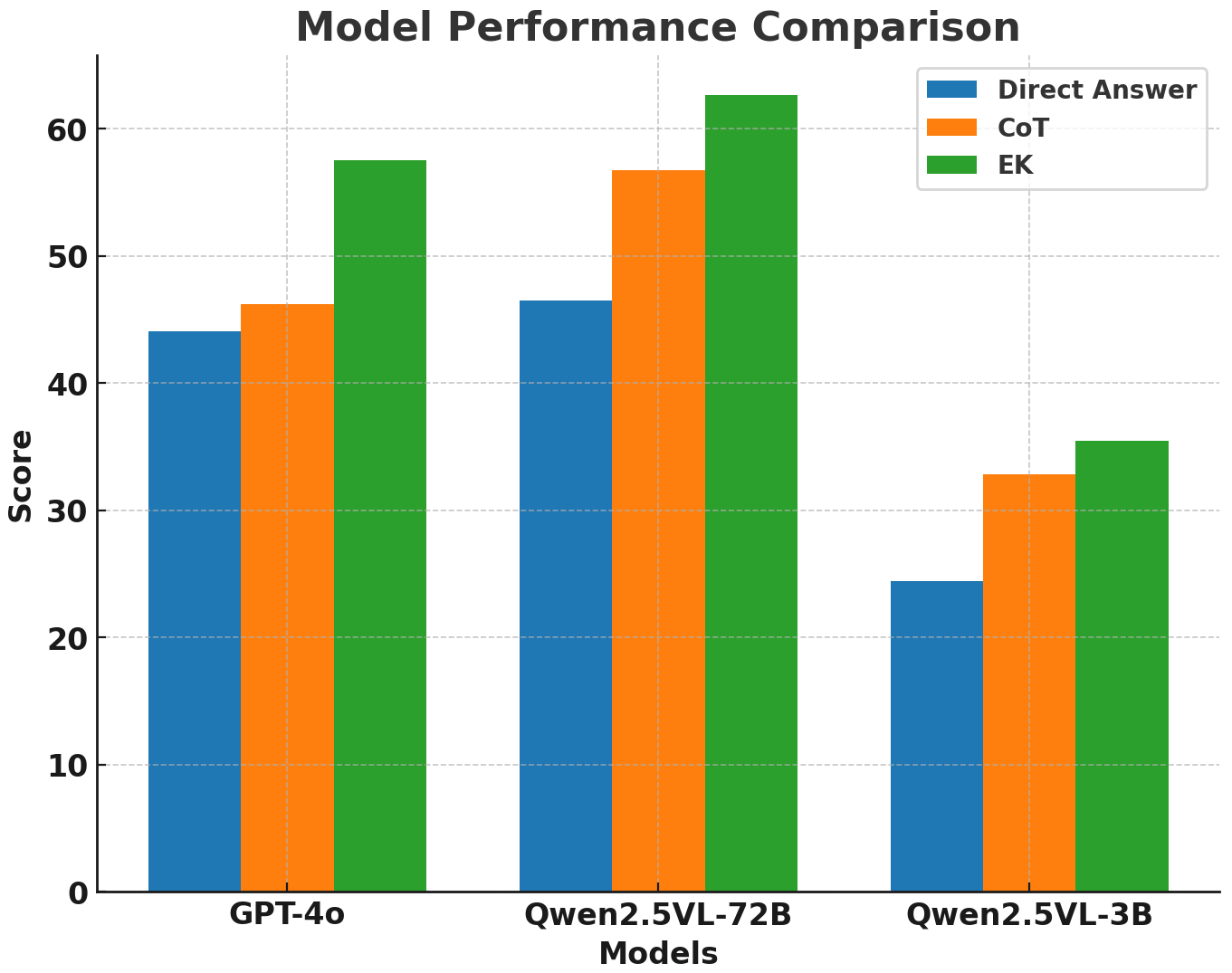}
    \vspace{-6pt}
    \caption{Localization performance comparison of three models with two enhancement techniques: Chain-of-Thought (CoT) and External Knowledge (EK).}
    \label{fig:counting_ablation}
    \vspace{-10pt}
\end{figure}

\noindent\textbf{Step 2a: Does Structured Perception Help?} To test whether perception is the bottleneck, we prompt models to first decompose the chemical diagram into a structured JSON representation before answering, following Chain-of-Thought methodologies~\cite{cot,sprague2024cot,yuwang,deng2024explicit}. As illustrated in Fig.~\ref{fig:counting_ablation}, this structured perception step yields only modest improvements for GPT-4o, Qwen2.5-VL-3B, and Qwen2.5-VL-72B (gains of 2.16\%, 8.38\%, and 10.30\%, respectively). While enforcing systematic visual analysis provides some benefit, the limited gains suggest the problem runs deeper than disorganized perception.

\noindent\textbf{Step 2b: What If Perception Were Perfect?} To eliminate the perception variable, we supply models with error-free structural information as External Knowledge (EK) alongside the image (see Appendix~\ref{app:ek}). This bypasses the models' visual processing, testing reasoning capabilities in a best-case scenario. Results in Fig.~\ref{fig:counting_ablation} are striking. While accuracy reaches 57.49\% for GPT-4o and 62.63\% for Qwen2.5-VL-72B, overall performance remains below 65\%. This confirms the primary bottleneck is not visual perception, but a deep-seated limitation in core reasoning over complex topologies.

\begin{table}[!t]
    \caption{Impact of molecular placeholder masking on model performance.}
    \vspace{-5pt}
    \centering
    \setlength{\tabcolsep}{2pt} 
    \small
    \begin{tabular}{lcccc}
    \toprule
     \textbf{Model} & \textbf{Task} & \makecell{\textbf{Original} \\ \textbf{images}} & \makecell{\textbf{Masked} \\ \textbf{images}} &  \makecell{\textbf{$\Delta$}} \\
    \midrule
    \multirow{4}{*}{\makecell{Qwen2.5-VL\\3B}}
        & Localization & 24.43 & 19.76 & -4.67 \\
        & Extraction & 89.61 & 78.50 & -11.11 \\
        & Tracing & 49.70 & 42.51 & -7.19 \\
        & Reasoning & 34.65 & 31.19 & -3.46 \\
    \cmidrule{1-5}
    \multirow{4}{*}{\makecell{Qwen2.5-VL\\7B}}
        & Localization & 37.96 & 35.81 & -2.15 \\
        & Extraction & 85.75 & 77.78 & -7.97 \\
        & Tracing & 66.47 & 56.89 & -9.58 \\
        & Reasoning & 39.11 & 47.52 & +8.41 \\
    \cmidrule{1-5}
    \multirow{4}{*}{\makecell{Qwen2.5-VL\\72B}}
        & Localization & 46.47 & 45.15 & -1.32 \\
        & Extraction & 89.86 & 80.68 & -9.18 \\
        & Tracing & 83.83 & 68.26 & -15.57 \\
        & Reasoning & 54.46 & 55.94 & +1.48 \\
    \bottomrule
    \end{tabular}
    \label{tab:ablation}
    \vspace{-10pt}
\end{table}

\subsection{How Reasoning Failures Manifest}

Having established that the bottleneck lies in reasoning rather than perception, we now examine how these failures manifest across the four task dimensions. 
The following cases highlight recurring failure patterns shared by different model families.

\noindent\textbf{Localization: Failure in Hierarchical Abstraction.} As shown in Fig.~\ref{fig:casestudy}(a), MLLMs correctly identify local textual markers, such as 1), 2), and 3), but cannot perform the hierarchical abstraction needed to understand these are conditions for a single global transformation. This leads to systematic over-segmentation when counting reaction steps, revealing an inability to move from local feature recognition to global process understanding.

\noindent\textbf{Extraction: Loss of Structural Consistency.} Tracking entities across a diagram requires maintaining focus throughout multiple visual elements. In Fig.~\ref{fig:casestudy}(b), GPT-4o correctly follows the main molecular transformation but is derailed by a local, irrelevant visual feature ($2HCl$) when identifying the final product subscript. This demonstrates a breakdown in structural consistency, where the model fails to filter distractors during extraction.

\noindent\textbf{Tracing: Failure in Topological Path Tracing.} MLLMs often disregard explicit topological rules in favor of learned heuristics. In Fig.~\ref{fig:casestudy}(c), models ignore directed arrows (5 $\rightarrow$ 4 $\rightarrow$ 3), predicting the next compound via superficial numerical sequences. This reveals a critical deficit in graph-theoretic reasoning and the inability to process connective logic.

\noindent\textbf{Reasoning: Inability to Classify Global Topology.} The ultimate test of structural reasoning is classifying a diagram's overall topology. In Fig.~\ref{fig:casestudy}(d), models correctly perceive local features like a linear sequence and a branching point, but fail to integrate these into the correct global classification of ``single line'' structure. This directly confirms our central finding: local perception is intact, but global structural reasoning fails.

\subsection{The Role of Molecular Structures in Topological Reasoning}

Having shown that reasoning is the main bottleneck, we further examine whether molecular structures serve as useful visual anchors or distractors for topology reasoning~\cite{li2025causally}. 
Following the masking procedure in Appendix~\ref{app:masking}, we replace molecular structures with black borders while preserving the original topological connectivity.

As shown in Tab.~\ref{tab:ablation}, masking molecular structures consistently reduces performance on Localization, Extraction, and Tracing across all model sizes. 
This suggests that models use molecular structures as visual anchors for organizing reaction connectivity and flow. 
In contrast, Reasoning shows a different pattern: the 3B model slightly declines (-3.46\%), whereas the 7B and 72B models improve by 8.41\% and 1.48\%. 
This suggests that molecular details help local tasks but may distract larger models from abstract topology reasoning.

Overall, molecular structures play a dual role: they support local perception and pathway understanding, but can impede the hierarchical abstraction required for global structural reasoning.

\vspace{-4pt}
\section{Conclusion}
\vspace{-4pt}
\label{sec:conclusion}
We introduce \ours, a benchmark to systematically investigate MLLMs' comprehension capabilities in chemical reaction diagrams. We evaluate MLLMs using 1,618 QA pairs based on real-world reaction images, covering tasks from basic recognition to advanced reasoning. Our findings reveal persistent limitations in MLLMs' ability to perform topological reasoning over chemical reaction diagrams, particularly when tasks require integrating local visual cues into a holistic understanding of reaction structure. This highlights the need for improved multimodal reasoning in scientific domains and provides a critical evaluation framework. 

\section*{Limitations}

Our benchmark focuses on chemical reaction diagrams rather than abstract graphs or other diagram types. This choice is deliberate. Chemical reaction diagrams provide natural topological diversity from linear chains to cyclic networks while demanding both local perception and global reasoning. Abstract synthetic graphs lack the visual complexity that makes findings practically meaningful, and other diagram types such as flowcharts typically exhibit limited topological variety. Extending this framework to other scientific domains remains valuable future work. Furthermore, although the scale of 1,618 instances is modest compared to massive, synthetically generated datasets, the exceptionally high density of multi-hop reasoning steps and the rigorous, multi-round expert annotation protocol ensure a high signal-to-noise ratio, effectively prioritizing analytical quality over sheer quantity.

\bibliography{custom}

@String(ECCV  = {ECCV})

@String(AAAI  = {AAAI})

@InProceedings{yuwang,
author="Wang, Yu
and Liu, Xiaogeng
and Li, Yu
and Chen, Muhao
and Xiao, Chaowei",
editor="Leonardis, Ale{\v{s}}
and Ricci, Elisa
and Roth, Stefan
and Russakovsky, Olga
and Sattler, Torsten
and Varol, G{\"u}l",
title="AdaShield : Safeguarding Multimodal Large Language Models from Structure-Based Attack via Adaptive Shield Prompting",
booktitle="Computer Vision -- ECCV 2024",
year="2025",
address="Cham",
pages="77--94",
}

@article{clip,
  author       = {Alec Radford and
                  Jong Wook Kim and
                  Chris Hallacy and
                  Aditya Ramesh and
                  Gabriel Goh and
                  Sandhini Agarwal and
                  Girish Sastry and
                  Amanda Askell and
                  Pamela Mishkin and
                  Jack Clark and
                  Gretchen Krueger and
                  Ilya Sutskever},
  title        = {Learning Transferable Visual Models From Natural Language Supervision},
  journal      = {CoRR},
  volume       = {abs/2103.00020},
  year         = {2021},
  url          = {https://arxiv.org/abs/2103.00020},
  eprinttype    = {arXiv},
  eprint       = {2103.00020},
  bibsource    = {dblp computer science bibliography, https://dblp.org}
}

@misc{li2023blip2,
      title={BLIP-2: Bootstrapping Language-Image Pre-training with Frozen Image Encoders and Large Language Models}, 
      author={Junnan Li and Dongxu Li and Silvio Savarese and Steven Hoi},
      year={2023},
      eprint={2301.12597},
      archivePrefix={arXiv},
      primaryClass={cs.CV},
      url={https://arxiv.org/abs/2301.12597}, 
}

@article{chartqa,
  title={Chartqa: A benchmark for question answering about charts with visual and logical reasoning},
  author={Masry, Ahmed and Long, Do Xuan and Tan, Jia Qing and Joty, Shafiq and Hoque, Enamul},
  journal={arXiv preprint arXiv:2203.10244},
  year={2022}
}

@inproceedings{docvqa,
  title={Docvqa: A dataset for vqa on document images},
  author={Mathew, Minesh and Karatzas, Dimosthenis and Jawahar, CV},
  booktitle={Proceedings of the IEEE/CVF winter conference on applications of computer vision},
  pages={2200--2209},
  year={2021}
}

@misc{qwen25vl,
      title={Qwen2.5-VL Technical Report}, 
      author={Shuai Bai and Keqin Chen and Xuejing Liu and Jialin Wang and Wenbin Ge and Sibo Song and Kai Dang and Peng Wang and Shijie Wang and Jun Tang and Humen Zhong and Yuanzhi Zhu and Mingkun Yang and Zhaohai Li and Jianqiang Wan and Pengfei Wang and Wei Ding and Zheren Fu and Yiheng Xu and Jiabo Ye and Xi Zhang and Tianbao Xie and Zesen Cheng and Hang Zhang and Zhibo Yang and Haiyang Xu and Junyang Lin},
      year={2025},
      eprint={2502.13923},
      archivePrefix={arXiv},
      primaryClass={cs.CV},
      url={https://arxiv.org/abs/2502.13923}, 
}

@article{internvl25,
  title={Expanding performance boundaries of open-source multimodal models with model, data, and test-time scaling},
  author={Chen, Zhe and Wang, Weiyun and Cao, Yue and Liu, Yangzhou and Gao, Zhangwei and Cui, Erfei and Zhu, Jinguo and Ye, Shenglong and Tian, Hao and Liu, Zhaoyang and others},
  journal={arXiv preprint arXiv:2412.05271},
  year={2024}
}

@misc{liu2024llavanext,
    title={LLaVA-NeXT: Improved reasoning, OCR, and world knowledge},
    url={https://llava-vl.github.io/blog/2024-01-30-llava-next/},
    author={Liu, Haotian and Li, Chunyuan and Li, Yuheng and Li, Bo and Zhang, Yuanhan and Shen, Sheng and Lee, Yong Jae},
    month={January},
    year={2024}
}

@article{minicpmo,
  title={MiniCPM-V: A GPT-4V Level MLLM on Your Phone},
  author={Yao, Yuan and Yu, Tianyu and Zhang, Ao and Wang, Chongyi and Cui, Junbo and Zhu, Hongji and Cai, Tianchi and Li, Haoyu and Zhao, Weilin and He, Zhihui and others},
  journal={arXiv preprint arXiv:2408.01800},
  year={2024}
}

@misc{deepseekvl2,
      title={DeepSeek-VL2: Mixture-of-Experts Vision-Language Models for Advanced Multimodal Understanding},
      author={Zhiyu Wu and Xiaokang Chen and Zizheng Pan and Xingchao Liu and Wen Liu and Damai Dai and Huazuo Gao and Yiyang Ma and Chengyue Wu and Bingxuan Wang and Zhenda Xie and Yu Wu and Kai Hu and Jiawei Wang and Yaofeng Sun and Yukun Li and Yishi Piao and Kang Guan and Aixin Liu and Xin Xie and Yuxiang You and Kai Dong and Xingkai Yu and Haowei Zhang and Liang Zhao and Yisong Wang and Chong Ruan},
      year={2024},
      eprint={2412.10302},
      archivePrefix={arXiv},
      primaryClass={cs.CV},
      url={https://arxiv.org/abs/2412.10302},
}

@article{phi35v,
  title={Phi-3 technical report: A highly capable language model locally on your phone},
  author={Abdin, Marah and Aneja, Jyoti and Awadalla, Hany and Awadallah, Ahmed and Awan, Ammar Ahmad and Bach, Nguyen and Bahree, Amit and Bakhtiari, Arash and Bao, Jianmin and Behl, Harkirat and others},
  journal={arXiv preprint arXiv:2404.14219},
  year={2024}
}

@misc{gpt4o,
      title={GPT-4o System Card}, 
      author={OpenAI},
      year={2024},
      eprint={2410.21276},
      archivePrefix={arXiv},
      primaryClass={cs.CL},
      url={https://arxiv.org/abs/2410.21276}, 
}

@misc{gpt55,
  title        = {Introducing GPT-5.5},
  author       = {{OpenAI}},
  year         = {2026},
  month        = apr,
  howpublished = {\url{https://openai.com/index/introducing-gpt-5-5/}}
}

@misc{claude35,
  author = {Anthropic},
  title = {Claude 3.5 Sonnet},
  year = {2024},
  url = {https://www.anthropic.com/news/claude-3-5-sonnet},
}

@misc{claudesonnet46,
  title        = {Introducing Claude Sonnet 4.6},
  author       = {{Anthropic}},
  year         = {2026},
  month        = feb,
  howpublished = {\url{https://www.anthropic.com/news/claude-sonnet-4-6}},
  note         = {Accessed: 2026-05-23}
}

@misc{gemini15,
      title={Gemini 1.5: Unlocking multimodal understanding across millions of tokens of context}, 
      author={{Gemini Team} and Petko Georgiev and Ving Ian Lei and Ryan Burnell and Libin Bai and Anmol Gulati and Garrett Tanzer and Damien Vincent and Zhufeng Pan and Shibo Wang and Soroosh Mariooryad and Yifan Ding and Xinyang Geng and Fred Alcober and Roy Frostig and Mark Omernick and Lexi Walker and Cosmin Paduraru and Christina Sorokin and Andrea Tacchetti and Colin Gaffney and Samira Daruki and Olcan Sercinoglu and Zach Gleicher and Juliette Love and Paul Voigtlaender and Rohan Jain and Gabriela Surita and Kareem Mohamed and Rory Blevins and Junwhan Ahn and Tao Zhu and Kornraphop Kawintiranon and Orhan Firat and Yiming Gu and Yujing Zhang and Matthew Rahtz and Manaal Faruqui and Natalie Clay and Justin Gilmer and JD Co-Reyes and Ivo Penchev and Rui Zhu and Nobuyuki Morioka and Kevin Hui and Krishna Haridasan and Victor Campos and Mahdis Mahdieh and Mandy Guo and Samer Hassan and Kevin Kilgour and Arpi Vezer and Heng-Tze Cheng and Raoul de Liedekerke and Siddharth Goyal and Paul Barham and DJ Strouse and Seb Noury and Jonas Adler and Mukund Sundararajan and Sharad Vikram and Dmitry Lepikhin and Michela Paganini and Xavier Garcia and Fan Yang and Dasha Valter and Maja Trebacz and Kiran Vodrahalli and Chulayuth Asawaroengchai and Roman Ring and Norbert Kalb and Livio Baldini Soares and Siddhartha Brahma and David Steiner and Tianhe Yu and Fabian Mentzer and Antoine He and Lucas Gonzalez and Bibo Xu and Raphael Lopez Kaufman and Laurent El Shafey and Junhyuk Oh and Tom Hennigan and George van den Driessche and Seth Odoom and Mario Lucic and Becca Roelofs and Sid Lall and Amit Marathe and Betty Chan and Santiago Ontanon and Luheng He and Denis Teplyashin and Jonathan Lai and Phil Crone and Bogdan Damoc and Lewis Ho and Sebastian Riedel and Karel Lenc and Chih-Kuan Yeh and Aakanksha Chowdhery and Yang Xu and Mehran Kazemi and Ehsan Amid and Anastasia Petrushkina and Kevin Swersky and Ali Khodaei and Gowoon Chen and Chris Larkin and Mario Pinto and Geng Yan and Adria Puigdomenech Badia and Piyush Patil and Steven Hansen and Dave Orr and Sebastien M. R. Arnold and Jordan Grimstad and Andrew Dai and Sholto Douglas and Rishika Sinha and Vikas Yadav and Xi Chen and Elena Gribovskaya and Jacob Austin and Jeffrey Zhao and Kaushal Patel and Paul Komarek and Sophia Austin and Sebastian Borgeaud and Linda Friso and Abhimanyu Goyal and Ben Caine and Kris Cao and Da-Woon Chung and Matthew Lamm and Gabe Barth-Maron and Thais Kagohara and Kate Olszewska and Mia Chen and Kaushik Shivakumar and Rishabh Agarwal and Harshal Godhia and Ravi Rajwar and Javier Snaider and Xerxes Dotiwalla and Yuan Liu and Aditya Barua and Victor Ungureanu and Yuan Zhang and Bat-Orgil Batsaikhan and Mateo Wirth and James Qin and Ivo Danihelka and Tulsee Doshi and Martin Chadwick and Jilin Chen and Sanil Jain and Quoc Le and Arjun Kar and Madhu Gurumurthy and Cheng Li and Ruoxin Sang and Fangyu Liu and Lampros Lamprou and Rich Munoz and Nathan Lintz and Harsh Mehta and Heidi Howard and Malcolm Reynolds and Lora Aroyo and Quan Wang and Lorenzo Blanco and Albin Cassirer and Jordan Griffith and Dipanjan Das and Stephan Lee and Jakub Sygnowski and Zach Fisher and James Besley and Richard Powell and Zafarali Ahmed and Dominik Paulus and David Reitter and Zalan Borsos and Rishabh Joshi and Aedan Pope and Steven Hand and Vittorio Selo and Vihan Jain and Nikhil Sethi and Megha Goel and Takaki Makino and Rhys May and Zhen Yang and Johan Schalkwyk and Christina Butterfield and Anja Hauth and Alex Goldin and Will Hawkins and Evan Senter and Sergey Brin and Oliver Woodman and Marvin Ritter and Eric Noland and Minh Giang and Vijay Bolina and Lisa Lee and Tim Blyth and Ian Mackinnon and Machel Reid and Obaid Sarvana and David Silver and Alexander Chen and Lily Wang and Loren Maggiore and Oscar Chang and Nithya Attaluri and Gregory Thornton and Chung-Cheng Chiu and Oskar Bunyan and Nir Levine and Timothy Chung and Evgenii Eltyshev and Xiance Si and Timothy Lillicrap and Demetra Brady and Vaibhav Aggarwal and Boxi Wu and Yuanzhong Xu and Ross McIlroy and Kartikeya Badola and Paramjit Sandhu and Erica Moreira and Wojciech Stokowiec and Ross Hemsley and Dong Li and Alex Tudor and Pranav Shyam and Elahe Rahimtoroghi and Salem Haykal and Pablo Sprechmann and Xiang Zhou and Diana Mincu and Yujia Li and Ravi Addanki and Kalpesh Krishna and Xiao Wu and Alexandre Frechette and Matan Eyal and Allan Dafoe and Dave Lacey and Jay Whang and Thi Avrahami and Ye Zhang and Emanuel Taropa and Hanzhao Lin and Daniel Toyama and Eliza Rutherford and Motoki Sano and HyunJeong Choe and Alex Tomala and Chalence Safranek-Shrader and Nora Kassner and Mantas Pajarskas and Matt Harvey and Sean Sechrist and Meire Fortunato and Christina Lyu and Gamaleldin Elsayed and Chenkai Kuang and James Lottes and Eric Chu and Chao Jia and Chih-Wei Chen and Peter Humphreys and Kate Baumli and Connie Tao and Rajkumar Samuel and Cicero Nogueira dos Santos and Anders Andreassen and Nemanja Rakićević and Dominik Grewe and Aviral Kumar and Stephanie Winkler and Jonathan Caton and Andrew Brock and Sid Dalmia and Hannah Sheahan and Iain Barr and Yingjie Miao and Paul Natsev and Jacob Devlin and Feryal Behbahani and Flavien Prost and Yanhua Sun and Artiom Myaskovsky and Thanumalayan Sankaranarayana Pillai and Dan Hurt and Angeliki Lazaridou and Xi Xiong and Ce Zheng and Fabio Pardo and Xiaowei Li and Dan Horgan and Joe Stanton and Moran Ambar and Fei Xia and Alejandro Lince and Mingqiu Wang and Basil Mustafa and Albert Webson and Hyo Lee and Rohan Anil and Martin Wicke and Timothy Dozat and Abhishek Sinha and Enrique Piqueras and Elahe Dabir and Shyam Upadhyay and Anudhyan Boral and Lisa Anne Hendricks and Corey Fry and Josip Djolonga and Yi Su and Jake Walker and Jane Labanowski and Ronny Huang and Vedant Misra and Jeremy Chen and RJ Skerry-Ryan and Avi Singh and Shruti Rijhwani and Dian Yu and Alex Castro-Ros and Beer Changpinyo and Romina Datta and Sumit Bagri and Arnar Mar Hrafnkelsson and Marcello Maggioni and Daniel Zheng and Yury Sulsky and Shaobo Hou and Tom Le Paine and Antoine Yang and Jason Riesa and Dominika Rogozinska and Dror Marcus and Dalia El Badawy and Qiao Zhang and Luyu Wang and Helen Miller and Jeremy Greer and Lars Lowe Sjos and Azade Nova and Heiga Zen and Rahma Chaabouni and Mihaela Rosca and Jiepu Jiang and Charlie Chen and Ruibo Liu and Tara Sainath and Maxim Krikun and Alex Polozov and Jean-Baptiste Lespiau and Josh Newlan and Zeyncep Cankara and Soo Kwak and Yunhan Xu and Phil Chen and Andy Coenen and Clemens Meyer and Katerina Tsihlas and Ada Ma and Juraj Gottweis and Jinwei Xing and Chenjie Gu and Jin Miao and Christian Frank and Zeynep Cankara and Sanjay Ganapathy and Ishita Dasgupta and Steph Hughes-Fitt and Heng Chen and David Reid and Keran Rong and Hongmin Fan and Joost van Amersfoort and Vincent Zhuang and Aaron Cohen and Shixiang Shane Gu and Anhad Mohananey and Anastasija Ilic and Taylor Tobin and John Wieting and Anna Bortsova and Phoebe Thacker and Emma Wang and Emily Caveness and Justin Chiu and Eren Sezener and Alex Kaskasoli and Steven Baker and Katie Millican and Mohamed Elhawaty and Kostas Aisopos and Carl Lebsack and Nathan Byrd and Hanjun Dai and Wenhao Jia and Matthew Wiethoff and Elnaz Davoodi and Albert Weston and Lakshman Yagati and Arun Ahuja and Isabel Gao and Golan Pundak and Susan Zhang and Michael Azzam and Khe Chai Sim and Sergi Caelles and James Keeling and Abhanshu Sharma and Andy Swing and YaGuang Li and Chenxi Liu and Carrie Grimes Bostock and Yamini Bansal and Zachary Nado and Ankesh Anand and Josh Lipschultz and Abhijit Karmarkar and Lev Proleev and Abe Ittycheriah and Soheil Hassas Yeganeh and George Polovets and Aleksandra Faust and Jiao Sun and Alban Rrustemi and Pen Li and Rakesh Shivanna and Jeremiah Liu and Chris Welty and Federico Lebron and Anirudh Baddepudi and Sebastian Krause and Emilio Parisotto and Radu Soricut and Zheng Xu and Dawn Bloxwich and Melvin Johnson and Behnam Neyshabur and Justin Mao-Jones and Renshen Wang and Vinay Ramasesh and Zaheer Abbas and Arthur Guez and Constant Segal and Duc Dung Nguyen and James Svensson and Le Hou and Sarah York and Kieran Milan and Sophie Bridgers and Wiktor Gworek and Marco Tagliasacchi and James Lee-Thorp and Michael Chang and Alexey Guseynov and Ale Jakse Hartman and Michael Kwong and Ruizhe Zhao and Sheleem Kashem and Elizabeth Cole and Antoine Miech and Richard Tanburn and Mary Phuong and Filip Pavetic and Sebastien Cevey and Ramona Comanescu and Richard Ives and Sherry Yang and Cosmo Du and Bo Li and Zizhao Zhang and Mariko Iinuma and Clara Huiyi Hu and Aurko Roy and Shaan Bijwadia and Zhenkai Zhu and Danilo Martins and Rachel Saputro and Anita Gergely and Steven Zheng and Dawei Jia and Ioannis Antonoglou and Adam Sadovsky and Shane Gu and Yingying Bi and Alek Andreev and Sina Samangooei and Mina Khan and Tomas Kocisky and Angelos Filos and Chintu Kumar and Colton Bishop and Adams Yu and Sarah Hodkinson and Sid Mittal and Premal Shah and Alexandre Moufarek and Yong Cheng and Adam Bloniarz and Jaehoon Lee and Pedram Pejman and Paul Michel and Stephen Spencer and Vladimir Feinberg and Xuehan Xiong and Nikolay Savinov and Charlotte Smith and Siamak Shakeri and Dustin Tran and Mary Chesus and Bernd Bohnet and George Tucker and Tamara von Glehn and Carrie Muir and Yiran Mao and Hideto Kazawa and Ambrose Slone and Kedar Soparkar and Disha Shrivastava and James Cobon-Kerr and Michael Sharman and Jay Pavagadhi and Carlos Araya and Karolis Misiunas and Nimesh Ghelani and Michael Laskin and David Barker and Qiujia Li and Anton Briukhov and Neil Houlsby and Mia Glaese and Balaji Lakshminarayanan and Nathan Schucher and Yunhao Tang and Eli Collins and Hyeontaek Lim and Fangxiaoyu Feng and Adria Recasens and Guangda Lai and Alberto Magni and Nicola De Cao and Aditya Siddhant and Zoe Ashwood and Jordi Orbay and Mostafa Dehghani and Jenny Brennan and Yifan He and Kelvin Xu and Yang Gao and Carl Saroufim and James Molloy and Xinyi Wu and Seb Arnold and Solomon Chang and Julian Schrittwieser and Elena Buchatskaya and Soroush Radpour and Martin Polacek and Skye Giordano and Ankur Bapna and Simon Tokumine and Vincent Hellendoorn and Thibault Sottiaux and Sarah Cogan and Aliaksei Severyn and Mohammad Saleh and Shantanu Thakoor and Laurent Shefey and Siyuan Qiao and Meenu Gaba and Shuo-yiin Chang and Craig Swanson and Biao Zhang and Benjamin Lee and Paul Kishan Rubenstein and Gan Song and Tom Kwiatkowski and Anna Koop and Ajay Kannan and David Kao and Parker Schuh and Axel Stjerngren and Golnaz Ghiasi and Gena Gibson and Luke Vilnis and Ye Yuan and Felipe Tiengo Ferreira and Aishwarya Kamath and Ted Klimenko and Ken Franko and Kefan Xiao and Indro Bhattacharya and Miteyan Patel and Rui Wang and Alex Morris and Robin Strudel and Vivek Sharma and Peter Choy and Sayed Hadi Hashemi and Jessica Landon and Mara Finkelstein and Priya Jhakra and Justin Frye and Megan Barnes and Matthew Mauger and Dennis Daun and Khuslen Baatarsukh and Matthew Tung and Wael Farhan and Henryk Michalewski and Fabio Viola and Felix de Chaumont Quitry and Charline Le Lan and Tom Hudson and Qingze Wang and Felix Fischer and Ivy Zheng and Elspeth White and Anca Dragan and Jean-baptiste Alayrac and Eric Ni and Alexander Pritzel and Adam Iwanicki and Michael Isard and Anna Bulanova and Lukas Zilka and Ethan Dyer and Devendra Sachan and Srivatsan Srinivasan and Hannah Muckenhirn and Honglong Cai and Amol Mandhane and Mukarram Tariq and Jack W. Rae and Gary Wang and Kareem Ayoub and Nicholas FitzGerald and Yao Zhao and Woohyun Han and Chris Alberti and Dan Garrette and Kashyap Krishnakumar and Mai Gimenez and Anselm Levskaya and Daniel Sohn and Josip Matak and Inaki Iturrate and Michael B. Chang and Jackie Xiang and Yuan Cao and Nishant Ranka and Geoff Brown and Adrian Hutter and Vahab Mirrokni and Nanxin Chen and Kaisheng Yao and Zoltan Egyed and Francois Galilee and Tyler Liechty and Praveen Kallakuri and Evan Palmer and Sanjay Ghemawat and Jasmine Liu and David Tao and Chloe Thornton and Tim Green and Mimi Jasarevic and Sharon Lin and Victor Cotruta and Yi-Xuan Tan and Noah Fiedel and Hongkun Yu and Ed Chi and Alexander Neitz and Jens Heitkaemper and Anu Sinha and Denny Zhou and Yi Sun and Charbel Kaed and Brice Hulse and Swaroop Mishra and Maria Georgaki and Sneha Kudugunta and Clement Farabet and Izhak Shafran and Daniel Vlasic and Anton Tsitsulin and Rajagopal Ananthanarayanan and Alen Carin and Guolong Su and Pei Sun and Shashank V and Gabriel Carvajal and Josef Broder and Iulia Comsa and Alena Repina and William Wong and Warren Weilun Chen and Peter Hawkins and Egor Filonov and Lucia Loher and Christoph Hirnschall and Weiyi Wang and Jingchen Ye and Andrea Burns and Hardie Cate and Diana Gage Wright and Federico Piccinini and Lei Zhang and Chu-Cheng Lin and Ionel Gog and Yana Kulizhskaya and Ashwin Sreevatsa and Shuang Song and Luis C. Cobo and Anand Iyer and Chetan Tekur and Guillermo Garrido and Zhuyun Xiao and Rupert Kemp and Huaixiu Steven Zheng and Hui Li and Ananth Agarwal and Christel Ngani and Kati Goshvadi and Rebeca Santamaria-Fernandez and Wojciech Fica and Xinyun Chen and Chris Gorgolewski and Sean Sun and Roopal Garg and Xinyu Ye and S. M. Ali Eslami and Nan Hua and Jon Simon and Pratik Joshi and Yelin Kim and Ian Tenney and Sahitya Potluri and Lam Nguyen Thiet and Quan Yuan and Florian Luisier and Alexandra Chronopoulou and Salvatore Scellato and Praveen Srinivasan and Minmin Chen and Vinod Koverkathu and Valentin Dalibard and Yaming Xu and Brennan Saeta and Keith Anderson and Thibault Sellam and Nick Fernando and Fantine Huot and Junehyuk Jung and Mani Varadarajan and Michael Quinn and Amit Raul and Maigo Le and Ruslan Habalov and Jon Clark and Komal Jalan and Kalesha Bullard and Achintya Singhal and Thang Luong and Boyu Wang and Sujeevan Rajayogam and Julian Eisenschlos and Johnson Jia and Daniel Finchelstein and Alex Yakubovich and Daniel Balle and Michael Fink and Sameer Agarwal and Jing Li and Dj Dvijotham and Shalini Pal and Kai Kang and Jaclyn Konzelmann and Jennifer Beattie and Olivier Dousse and Diane Wu and Remi Crocker and Chen Elkind and Siddhartha Reddy Jonnalagadda and Jong Lee and Dan Holtmann-Rice and Krystal Kallarackal and Rosanne Liu and Denis Vnukov and Neera Vats and Luca Invernizzi and Mohsen Jafari and Huanjie Zhou and Lilly Taylor and Jennifer Prendki and Marcus Wu and Tom Eccles and Tianqi Liu and Kavya Kopparapu and Francoise Beaufays and Christof Angermueller and Andreea Marzoca and Shourya Sarcar and Hilal Dib and Jeff Stanway and Frank Perbet and Nejc Trdin and Rachel Sterneck and Andrey Khorlin and Dinghua Li and Xihui Wu and Sonam Goenka and David Madras and Sasha Goldshtein and Willi Gierke and Tong Zhou and Yaxin Liu and Yannie Liang and Anais White and Yunjie Li and Shreya Singh and Sanaz Bahargam and Mark Epstein and Sujoy Basu and Li Lao and Adnan Ozturel and Carl Crous and Alex Zhai and Han Lu and Zora Tung and Neeraj Gaur and Alanna Walton and Lucas Dixon and Ming Zhang and Amir Globerson and Grant Uy and Andrew Bolt and Olivia Wiles and Milad Nasr and Ilia Shumailov and Marco Selvi and Francesco Piccinno and Ricardo Aguilar and Sara McCarthy and Misha Khalman and Mrinal Shukla and Vlado Galic and John Carpenter and Kevin Villela and Haibin Zhang and Harry Richardson and James Martens and Matko Bosnjak and Shreyas Rammohan Belle and Jeff Seibert and Mahmoud Alnahlawi and Brian McWilliams and Sankalp Singh and Annie Louis and Wen Ding and Dan Popovici and Lenin Simicich and Laura Knight and Pulkit Mehta and Nishesh Gupta and Chongyang Shi and Saaber Fatehi and Jovana Mitrovic and Alex Grills and Joseph Pagadora and Tsendsuren Munkhdalai and Dessie Petrova and Danielle Eisenbud and Zhishuai Zhang and Damion Yates and Bhavishya Mittal and Nilesh Tripuraneni and Yannis Assael and Thomas Brovelli and Prateek Jain and Mihajlo Velimirovic and Canfer Akbulut and Jiaqi Mu and Wolfgang Macherey and Ravin Kumar and Jun Xu and Haroon Qureshi and Gheorghe Comanici and Jeremy Wiesner and Zhitao Gong and Anton Ruddock and Matthias Bauer and Nick Felt and Anirudh GP and Anurag Arnab and Dustin Zelle and Jonas Rothfuss and Bill Rosgen and Ashish Shenoy and Bryan Seybold and Xinjian Li and Jayaram Mudigonda and Goker Erdogan and Jiawei Xia and Jiri Simsa and Andrea Michi and Yi Yao and Christopher Yew and Steven Kan and Isaac Caswell and Carey Radebaugh and Andre Elisseeff and Pedro Valenzuela and Kay McKinney and Kim Paterson and Albert Cui and Eri Latorre-Chimoto and Solomon Kim and William Zeng and Ken Durden and Priya Ponnapalli and Tiberiu Sosea and Christopher A. Choquette-Choo and James Manyika and Brona Robenek and Harsha Vashisht and Sebastien Pereira and Hoi Lam and Marko Velic and Denese Owusu-Afriyie and Katherine Lee and Tolga Bolukbasi and Alicia Parrish and Shawn Lu and Jane Park and Balaji Venkatraman and Alice Talbert and Lambert Rosique and Yuchung Cheng and Andrei Sozanschi and Adam Paszke and Praveen Kumar and Jessica Austin and Lu Li and Khalid Salama and Bartek Perz and Wooyeol Kim and Nandita Dukkipati and Anthony Baryshnikov and Christos Kaplanis and XiangHai Sheng and Yuri Chervonyi and Caglar Unlu and Diego de Las Casas and Harry Askham and Kathryn Tunyasuvunakool and Felix Gimeno and Siim Poder and Chester Kwak and Matt Miecnikowski and Vahab Mirrokni and Alek Dimitriev and Aaron Parisi and Dangyi Liu and Tomy Tsai and Toby Shevlane and Christina Kouridi and Drew Garmon and Adrian Goedeckemeyer and Adam R. Brown and Anitha Vijayakumar and Ali Elqursh and Sadegh Jazayeri and Jin Huang and Sara Mc Carthy and Jay Hoover and Lucy Kim and Sandeep Kumar and Wei Chen and Courtney Biles and Garrett Bingham and Evan Rosen and Lisa Wang and Qijun Tan and David Engel and Francesco Pongetti and Dario de Cesare and Dongseong Hwang and Lily Yu and Jennifer Pullman and Srini Narayanan and Kyle Levin and Siddharth Gopal and Megan Li and Asaf Aharoni and Trieu Trinh and Jessica Lo and Norman Casagrande and Roopali Vij and Loic Matthey and Bramandia Ramadhana and Austin Matthews and CJ Carey and Matthew Johnson and Kremena Goranova and Rohin Shah and Shereen Ashraf and Kingshuk Dasgupta and Rasmus Larsen and Yicheng Wang and Manish Reddy Vuyyuru and Chong Jiang and Joana Ijazi and Kazuki Osawa and Celine Smith and Ramya Sree Boppana and Taylan Bilal and Yuma Koizumi and Ying Xu and Yasemin Altun and Nir Shabat and Ben Bariach and Alex Korchemniy and Kiam Choo and Olaf Ronneberger and Chimezie Iwuanyanwu and Shubin Zhao and David Soergel and Cho-Jui Hsieh and Irene Cai and Shariq Iqbal and Martin Sundermeyer and Zhe Chen and Elie Bursztein and Chaitanya Malaviya and Fadi Biadsy and Prakash Shroff and Inderjit Dhillon and Tejasi Latkar and Chris Dyer and Hannah Forbes and Massimo Nicosia and Vitaly Nikolaev and Somer Greene and Marin Georgiev and Pidong Wang and Nina Martin and Hanie Sedghi and John Zhang and Praseem Banzal and Doug Fritz and Vikram Rao and Xuezhi Wang and Jiageng Zhang and Viorica Patraucean and Dayou Du and Igor Mordatch and Ivan Jurin and Lewis Liu and Ayush Dubey and Abhi Mohan and Janek Nowakowski and Vlad-Doru Ion and Nan Wei and Reiko Tojo and Maria Abi Raad and Drew A. Hudson and Vaishakh Keshava and Shubham Agrawal and Kevin Ramirez and Zhichun Wu and Hoang Nguyen and Ji Liu and Madhavi Sewak and Bryce Petrini and DongHyun Choi and Ivan Philips and Ziyue Wang and Ioana Bica and Ankush Garg and Jarek Wilkiewicz and Priyanka Agrawal and Xiaowei Li and Danhao Guo and Emily Xue and Naseer Shaik and Andrew Leach and Sadh MNM Khan and Julia Wiesinger and Sammy Jerome and Abhishek Chakladar and Alek Wenjiao Wang and Tina Ornduff and Folake Abu and Alireza Ghaffarkhah and Marcus Wainwright and Mario Cortes and Frederick Liu and Joshua Maynez and Andreas Terzis and Pouya Samangouei and Riham Mansour and Tomasz Kępa and François-Xavier Aubet and Anton Algymr and Dan Banica and Agoston Weisz and Andras Orban and Alexandre Senges and Ewa Andrejczuk and Mark Geller and Niccolo Dal Santo and Valentin Anklin and Majd Al Merey and Martin Baeuml and Trevor Strohman and Junwen Bai and Slav Petrov and Yonghui Wu and Demis Hassabis and Koray Kavukcuoglu and Jeff Dean and Oriol Vinyals},
      year={2024},
      eprint={2403.05530},
      archivePrefix={arXiv},
      primaryClass={cs.CL},
      url={https://arxiv.org/abs/2403.05530}, 
}

@misc{gemini31pro,
  title        = {Gemini 3.1 Pro Model Card},
  author       = {{Google DeepMind}},
  year         = {2026},
  month        = feb,
  howpublished = {\url{https://deepmind.google/models/model-cards/gemini-3-1-pro/}},
  note         = {Accessed: 2026-05-23}
}

@misc{qwenvlmax,
  author = {Qwen},
  title = {Qwen-VL-Max},
  year = {2024},
  url = {https://huggingface.co/spaces/Qwen/Qwen-VL-Max},
}

@article{rxnscribe,
  title={RxnScribe: a sequence generation model for reaction diagram parsing},
  author={Qian, Yujie and Guo, Jiang and Tu, Zhengkai and Coley, Connor W and Barzilay, Regina},
  journal={Journal of chemical information and modeling},
  volume={63},
  number={13},
  pages={4030--4041},
  year={2023},
  publisher={ACS Publications}
}

@article{cot,
  title={Chain-of-thought prompting elicits reasoning in large language models},
  author={Wei, Jason and Wang, Xuezhi and Schuurmans, Dale and Bosma, Maarten and Xia, Fei and Chi, Ed and Le, Quoc V and Zhou, Denny and others},
  journal={Advances in neural information processing systems},
  volume={35},
  pages={24824--24837},
  year={2022}
}

@article{molscribe,
  title={MolScribe: robust molecular structure recognition with image-to-graph generation},
  author={Qian, Yujie and Guo, Jiang and Tu, Zhengkai and Li, Zhening and Coley, Connor W and Barzilay, Regina},
  journal={Journal of chemical information and modeling},
  volume={63},
  number={7},
  pages={1925--1934},
  year={2023},
  publisher={ACS Publications}
}

@article{wilary2023reactiondataextractor,
  title={ReactionDataExtractor 2.0: a deep learning approach for data extraction from chemical reaction schemes},
  author={Wilary, Damian M and Cole, Jacqueline M},
  journal={Journal of Chemical Information and Modeling},
  volume={63},
  number={19},
  pages={6053--6067},
  year={2023},
  publisher={ACS Publications}
}

@article{sprague2024cot,
  title={To cot or not to cot? chain-of-thought helps mainly on math and symbolic reasoning},
  author={Sprague, Zayne and Yin, Fangcong and Rodriguez, Juan Diego and Jiang, Dongwei and Wadhwa, Manya and Singhal, Prasann and Zhao, Xinyu and Ye, Xi and Mahowald, Kyle and Durrett, Greg},
  journal={arXiv preprint arXiv:2409.12183},
  year={2024}
}

@article{deng2024explicit,
  title={From explicit cot to implicit cot: Learning to internalize cot step by step},
  author={Deng, Yuntian and Choi, Yejin and Shieber, Stuart},
  journal={arXiv preprint arXiv:2405.14838},
  year={2024}
}

@inproceedings{chen2024mllm,
  title={Mllm-as-a-judge: Assessing multimodal llm-as-a-judge with vision-language benchmark},
  author={Chen, Dongping and Chen, Ruoxi and Zhang, Shilin and Wang, Yaochen and Liu, Yinuo and Zhou, Huichi and Zhang, Qihui and Wan, Yao and Zhou, Pan and Sun, Lichao},
  booktitle={Forty-first International Conference on Machine Learning},
  year={2024}
}

@inproceedings{textvqa,
  title={Towards vqa models that can read},
  author={Singh, Amanpreet and Natarajan, Vivek and Shah, Meet and Jiang, Yu and Chen, Xinlei and Batra, Dhruv and Parikh, Devi and Rohrbach, Marcus},
  booktitle={Proceedings of the IEEE/CVF conference on computer vision and pattern recognition},
  pages={8317--8326},
  year={2019}
}

@article{scienceqa,
  title={Learn to explain: Multimodal reasoning via thought chains for science question answering},
  author={Lu, Pan and Mishra, Swaroop and Xia, Tanglin and Qiu, Liang and Chang, Kai-Wei and Zhu, Song-Chun and Tafjord, Oyvind and Clark, Peter and Kalyan, Ashwin},
  journal={Advances in Neural Information Processing Systems},
  volume={35},
  pages={2507--2521},
  year={2022}
}

@article{decimer,
  title={DECIMER: towards deep learning for chemical image recognition},
  author={Rajan, Kohulan and Zielesny, Achim and Steinbeck, Christoph},
  journal={Journal of Cheminformatics},
  volume={12},
  number={1},
  pages={65},
  year={2020},
  publisher={Springer}
}

@article{img2mol,
  title={Img2Mol--accurate SMILES recognition from molecular graphical depictions},
  author={Clevert, Djork-Arn{\'e} and Le, Tuan and Winter, Robin and Montanari, Floriane},
  journal={Chemical science},
  volume={12},
  number={42},
  pages={14174--14181},
  year={2021},
  publisher={Royal Society of Chemistry}
}

@misc{optical,
  title={Optical structure recognition software to recover chemical information: OSRA, an open source solution},
  author={Filippov, Igor V and Nicklaus, Marc C},
  year={2009},
  publisher={ACS Publications}
}

@article{chemgrapher,
  title={ChemGrapher: optical graph recognition of chemical compounds by deep learning},
  author={Oldenhof, Martijn and Arany, Adam and Moreau, Yves and Simm, Jaak},
  journal={Journal of chemical information and modeling},
  volume={60},
  number={10},
  pages={4506--4517},
  year={2020},
  publisher={ACS Publications}
}

@article{decimer_ai,
  title={DECIMER. ai: an open platform for automated optical chemical structure identification, segmentation and recognition in scientific publications},
  author={Rajan, Kohulan and Brinkhaus, Henning Otto and Agea, M Isabel and Zielesny, Achim and Steinbeck, Christoph},
  journal={Nature communications},
  volume={14},
  number={1},
  pages={5045},
  year={2023},
  publisher={Nature Publishing Group UK London}
}

@article{li2025preference,
  title={Preference leakage: A contamination problem in llm-as-a-judge},
  author={Li, Dawei and Sun, Renliang and Huang, Yue and Zhong, Ming and Jiang, Bohan and Han, Jiawei and Zhang, Xiangliang and Wang, Wei and Liu, Huan},
  journal={arXiv preprint arXiv:2502.01534},
  year={2025}
}

@article{gemini2.5,
  title={Gemini 2.5: Pushing the frontier with advanced reasoning, multimodality, long context, and next generation agentic capabilities},
  author={Comanici, Gheorghe and Bieber, Eric and Schaekermann, Mike and Pasupat, Ice and Sachdeva, Noveen and Dhillon, Inderjit and Blistein, Marcel and Ram, Ori and Zhang, Dan and Rosen, Evan and others},
  journal={arXiv preprint arXiv:2507.06261},
  year={2025}
}

@inproceedings{ai2d,
  title={A diagram is worth a dozen images},
  author={Kembhavi, Aniruddha and Salvato, Mike and Kolve, Eric and Seo, Minjoon and Hajishirzi, Hannaneh and Farhadi, Ali},
  booktitle={European conference on computer vision},
  pages={235--251},
  year={2016},
  organization={Springer}
}

@article{qwen3vl,
  title={Qwen3-vl technical report},
  author={Bai, Shuai and Cai, Yuxuan and Chen, Ruizhe and Chen, Keqin and Chen, Xionghui and Cheng, Zesen and Deng, Lianghao and Ding, Wei and Gao, Chang and Ge, Chunjiang and others},
  journal={arXiv preprint arXiv:2511.21631},
  year={2025}
}

@misc{gemma4,
  author = {{Google DeepMind}},
  title = {Gemma 4},
  year = {2026},
  url = {https://deepmind.google/models/gemma/gemma-4/},
  note = {Accessed: 2026-05-23}
}

@misc{kimi-k2.6,
  author = {{Moonshot AI}},
  title = {Kimi K2.6},
  year = {2026},
  url = {https://www.kimi.com/blog/kimi-k2-6},
  note = {Accessed: 2026-05-23}
}

@article{chemfp,
  title={The chemfp project},
  author={Dalke, Andrew},
  journal={Journal of cheminformatics},
  volume={11},
  number={1},
  pages={76},
  year={2019},
  publisher={Springer}
}

@article{chemo,
  title={ChemLabs on ChemO: a multi-agent system for multimodal reasoning on IChO 2025},
  author={Xu, Qiang and Bai, Shengyuan and Chen, Leqing and Liu, Zijing and Li, Yu},
  journal={arXiv preprint arXiv:2511.16205},
  year={2025}
}

@article{mozi,
  title={Mozi: Governed autonomy for drug discovery llm agents},
  author={Cao, He and Liu, Siyu and Zhang, Fan and Liu, Zijing and Li, Hao and Feng, Bin and Bai, Shengyuan and Chen, Leqing and Xie, Kai and Li, Yu},
  journal={arXiv preprint arXiv:2603.03655},
  year={2026}
}

@inproceedings{counterfactual,
  title={Counterfactual-based cognitive alignment in-context learning for relation extraction},
  author={Li, Qibin and Bai, Shengyuan and Zhou, Nai and Yao, Nianmin},
  booktitle={Proceedings of the AAAI Conference on Artificial Intelligence},
  volume={40},
  number={27},
  pages={23087--23095},
  year={2026}
}

@inproceedings{bai2025enhancing,
  title={Enhancing nlu in large language models using adversarial noisy instruction tuning},
  author={Bai, Shengyuan and Li, Qibin and Wang, Zhe and Zhou, Nai and Yao, Nianmin},
  booktitle={Proceedings of the AAAI conference on artificial intelligence},
  volume={39},
  number={22},
  pages={23451--23459},
  year={2025}
}

@article{li2025causally,
  title={Causally graph-guided counterfactual analysis to biomedical named entity recognition},
  author={Li, Qibin and Bai, Shengyuan and Zhou, Nai and Yao, Nianmin},
  journal={Expert Systems with Applications},
  pages={130485},
  year={2025},
  publisher={Elsevier}
}

@article{li2025enhancing,
  title={Enhancing entity and relation extraction with dynamic hard negative augmentation framework},
  author={Li, Qibin and Bai, Shengyuan and Zhou, Nai and Yao, Nianmin},
  journal={Engineering Applications of Artificial Intelligence},
  volume={161},
  pages={112211},
  year={2025},
  publisher={Elsevier}
}

@inproceedings{moleculeqa,
  title={Moleculeqa: A dataset to evaluate factual accuracy in molecular comprehension},
  author={Lu, Xingyu and Cao, He and Liu, Zijing and Bai, Shengyuan and Chen, Leqing and Yao, Yuan and Zheng, Hai-Tao and Li, Yu},
  booktitle={Findings of the Association for Computational Linguistics: EMNLP 2024},
  pages={3769--3789},
  year={2024}
}

@inproceedings{
beyond,
title={Beyond Chemical {QA}: Evaluating {LLM}'s Chemical Reasoning with Modular Chemical Operations},
author={Hao Li and He Cao and Bin Feng and Yanjun Shao and Xiangru Tang and Zhiyuan Yan and Yonghong Tian and Li Yuan and Yu Li},
booktitle={The Thirty-ninth Annual Conference on Neural Information Processing Systems Datasets and Benchmarks Track},
year={2026},
}

\clearpage

\appendix

\section{Robustness of the Evaluation Protocol}

To ensure our findings are not artifacts of an overly strict evaluation metric, we compare exact-match scoring against an LLM-as-a-judge approach. Strict exact-match risks penalizing semantically correct answers due to surface-form variations (e.g., ``2'' vs. ``two'').

We explicitly mitigate such variations in ReactBench through constrained task design, including single-character multiple-choice questions and strict prompts for numeric outputs. To quantify potential underestimation, we relax these constraints and use \texttt{Gemini-2.5-Flash}~\cite{gemini2.5} as an external semantic judge.

As shown in Tab.~\ref{tab:eval_comparison}, semantic evaluation changes the absolute scores substantially (e.g., Tracing improves from 49.70\% to 70.06\% for Qwen2.5-VL-3B). Despite some task-level variations, the overall model ranking remains consistent, and the significant capability gaps persist. This indicates that while the absolute scale shifts, the overall comparative assessment of model capabilities remains stable.

We therefore retain exact-match as the primary metric for ReactBench. It guarantees absolute reproducibility, eliminates the computational overhead and inherent biases of LLM judges~\cite{li2025preference}, and provides a rigorous lower-bound standard for future benchmarking.

\section{Question-Only Blind Baseline}
\label{app:blind_baseline}

A natural concern for any diagram-based benchmark is whether models can
exploit linguistic patterns or dataset priors to guess answers without
actually reading the diagram. To rule out this shortcut, we conduct a
\emph{question-only} (blind) ablation in which the chemical reaction
diagram is removed entirely and the model receives the textual question
alone. We evaluate the strongest open-source family in our study,
Qwen3-VL-Instruct (8B and 32B), under identical prompts and answer
extraction rules as the main experiments.

\begin{table}[!t]
\caption{Comparison of evaluation protocols for Qwen2.5-VL. \textbf{EM}: Exact Match; \textbf{Judge}: LLM-as-a-Judge; \textbf{Loc.}: Localization; \textbf{Ext.}: Extraction; \textbf{Trac.}: Tracing; \textbf{Reas.}: Reasoning; \textbf{Avg.}: arithmetic mean over the four tasks.}
\label{tab:eval_comparison}
\centering
\small
\setlength{\tabcolsep}{3.2pt}
\begin{tabular}{llccccc}
\toprule
\textbf{Size} & \textbf{Protocol} & \textbf{Loc.} & \textbf{Ext.} & \textbf{Trac.} & \textbf{Reas.} & \textbf{Avg.} \\
\midrule
\multirow{2}{*}{3B} 
& EM    & 24.43 & 89.61 & 49.70 & 34.65 & 49.60 \\
& Judge & 36.40 & 83.33 & 70.06 & 25.25 & 53.76 \\
\midrule
\multirow{2}{*}{7B} 
& EM    & 37.96 & 85.75 & 66.47 & 39.11 & 57.32 \\
& Judge & 54.01 & 87.44 & 87.42 & 42.57 & 67.86 \\
\bottomrule
\end{tabular}
\end{table}

\begin{table*}[t]
\centering
\small
\setlength{\tabcolsep}{5pt}
\begin{tabular}{llcccc}
\toprule
\textbf{Model} & \textbf{Input} & \textbf{Loc.} & \textbf{Ext.} & \textbf{Trac.} & \textbf{Reas.} \\
\midrule
\multirow{2}{*}{Qwen3-VL-Instruct-8B}
 & Image + Question & 52.34 & 85.02 & 82.04 & 32.67 \\
 & Question Only    & 15.57 &  6.76 & 36.53 &  9.90 \\
\midrule
\multirow{2}{*}{Qwen3-VL-Instruct-32B}
 & Image + Question & 60.24 & 86.96 & 91.02 & 57.43 \\
 & Question Only    & 10.30 &  5.56 & 32.93 &  8.91 \\
\bottomrule
\end{tabular}
\caption{Question-only (blind) baseline on ReactBench. Removing the
reaction diagram causes performance to collapse across all four
dimensions. Loc.: localization; Ext.: extraction; Trac.: tracing;
Reas.: reasoning.}
\label{tab:blind_baseline}
\end{table*}

\noindent\textbf{Performance collapses without visual input.}
As shown in Tab.~\ref{tab:blind_baseline}, removing the diagram degrades
performance across all four dimensions. Extraction, which depends on
reading numerical attributes directly off the diagram, drops to below
$7\%$ for both model scales, and Reasoning falls below $10\%$. Notably,
these scores fall \emph{below} the random-guessing level of the
corresponding answer spaces, indicating that blind models do not guess
uniformly but instead commit to systematically incorrect default answers
(e.g., a canonical endpoint count). This confirms that ReactBench
contains no exploitable language prior: the questions are not answerable
from templated phrasing or from the marginal distribution of ground-truth
answers.

\noindent\textbf{Visual topology is required, not optional.}
Questions such as \textit{``What is the subscript of the immediate next
compound?''} cannot be resolved by heuristics, since the answer depends
on multi-hop traversal over the specific topology of each diagram. Because
our 1{,}300+ diagrams span linear chains, branched mechanisms, trees, and
cyclic graphs, no single structural prior generalizes across instances.
The residual accuracy observed on Tracing ($32.93$--$36.53\%$) reflects
the fact that a subset of tracing questions admits a small candidate set
of adjacent subscripts, rather than any genuine reasoning ability.

\paragraph{Corroboration from thinking-style models.}
We repeat the same blind experiment on the reasoning-focused variants
(Qwen3-VL-Thinking 8B/32B). Because these models perform explicit
chain-of-thought before answering, they consistently detect that no
diagram is available and decline to produce an answer rather than
hallucinating one. This behavior further indicates that ReactBench
questions are strictly grounded in the visual input and are not solvable
from the text alone.

\section{Empirical Validation of OCSR Limitations}
\label{sec:appendix_ocsr_limitations}

To illustrate the information loss inherent in current OCSR tools, we analyze the output of RxnScribe~\cite{rxnscribe}, a state-of-the-art method. The following example displays a typical prediction for a reaction scheme:

\begin{json}
[
  {
    "reactants": [
      {
        "category": "[Mol]",
        "smiles": "Cc1ncc(C[n+]2csc(CCO)c2C)c(N)n1.Cl"
      }
    ],
    "conditions": [
      { "category": "[Txt]", "text": "28.5% aq" },
      { "category": "[Mol]", "smiles": "O=S(=O)([O][Na])SCC1CCCO1" },
      { "category": "[Txt]", "text": "NaOH" },
      { "category": "[Txt]", "text": "rt, 10 min" }
    ],
    "products": [
      {
        "category": "[Mol]",
        "smiles": "C/C(=C(\\CCO)SSCC1CCCO1)N(C=O)Cc1cnc(C)nc1N"
      }
    ]
  }
]
\end{json}

While this representation correctly recovers individual molecules and textual labels, it fails to encode the reaction topology targeted by our benchmark:
\begin{itemize}
    \item \textbf{Connectivity:} It does not specify which arrow connects which reactant to which product.
    \item \textbf{Sequential Order:} It does not indicate the temporal order of individual steps in a multi-step scheme.
    \item \textbf{Global Structure:} It does not represent branching pathways or convergent steps within the global network.
\end{itemize}
Consequently, a model relying solely on OCSR-derived textual representations (SMILES or JSON) lacks the necessary structural context to perform complex topological reasoning.

\section{Detailed Experiment Setups}
\label{sec:detailed_setup}
In this section, we provide more details about our experiment designs.

\subsection{Illustration of Ground-Truth Structured JSON}
\label{app:gt_json}

\begin{figure*}[!ht]
    \vspace{-20pt}
    \centering    
    \includegraphics[width=0.88\linewidth]{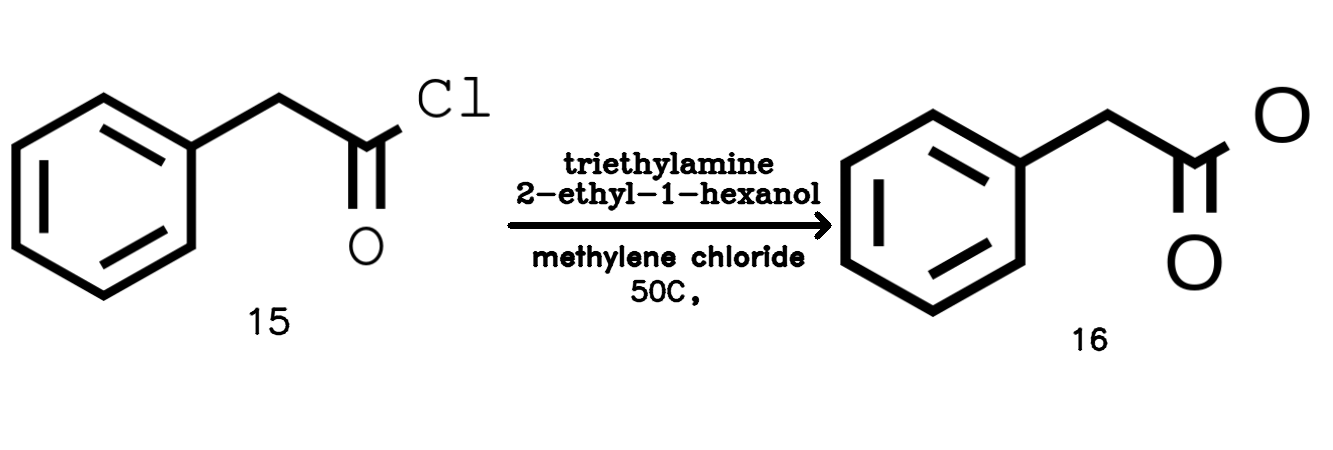}
    \vspace{-20pt}
    \caption{\textbf{Example of a chemical reaction diagram used for the text-only ablation study.} This visual representation is translated into the structured JSON format shown below, which preserves the chemical semantics (SMILES, text).}
    \label{fig:gt_json_example}
\end{figure*}

In this section, we provide a concrete example of the ground-truth structured JSON used in the text-only ablation study discussed in the main paper. Fig.~\ref{fig:gt_json_example} shows the original chemical reaction diagram, while the subsequent JSON snippet demonstrates how its constituent elements are computationally structured. 

\begin{gt}
[
  {
    "reactants": [
      {
        "category": "[Mol]",
        "smiles": "ClC(CC1=CC=CC=C1)=O"
      }
    ],
    "conditions": [
      { "category": "[Txt]", "text": "triethylamine" },
      { "category": "[Txt]", "text": "2-ethyl-1-hexanol" },
      { "category": "[Txt]", "text": "methylene chloride" },
      { "category": "[Txt]", "text": "50C" }
    ],
    "products": [
      {
        "category": "[Mol]",
        "smiles": "[O]C(CC1=CC=CC=C1)=O"
      }
    ]
  }
]
\end{gt}




\subsection{Prompt for Answer Extraction}
\label{app:prompt}
\begin{figure*}[!ht] 
    \centering    
    \includegraphics[width=0.88\linewidth]{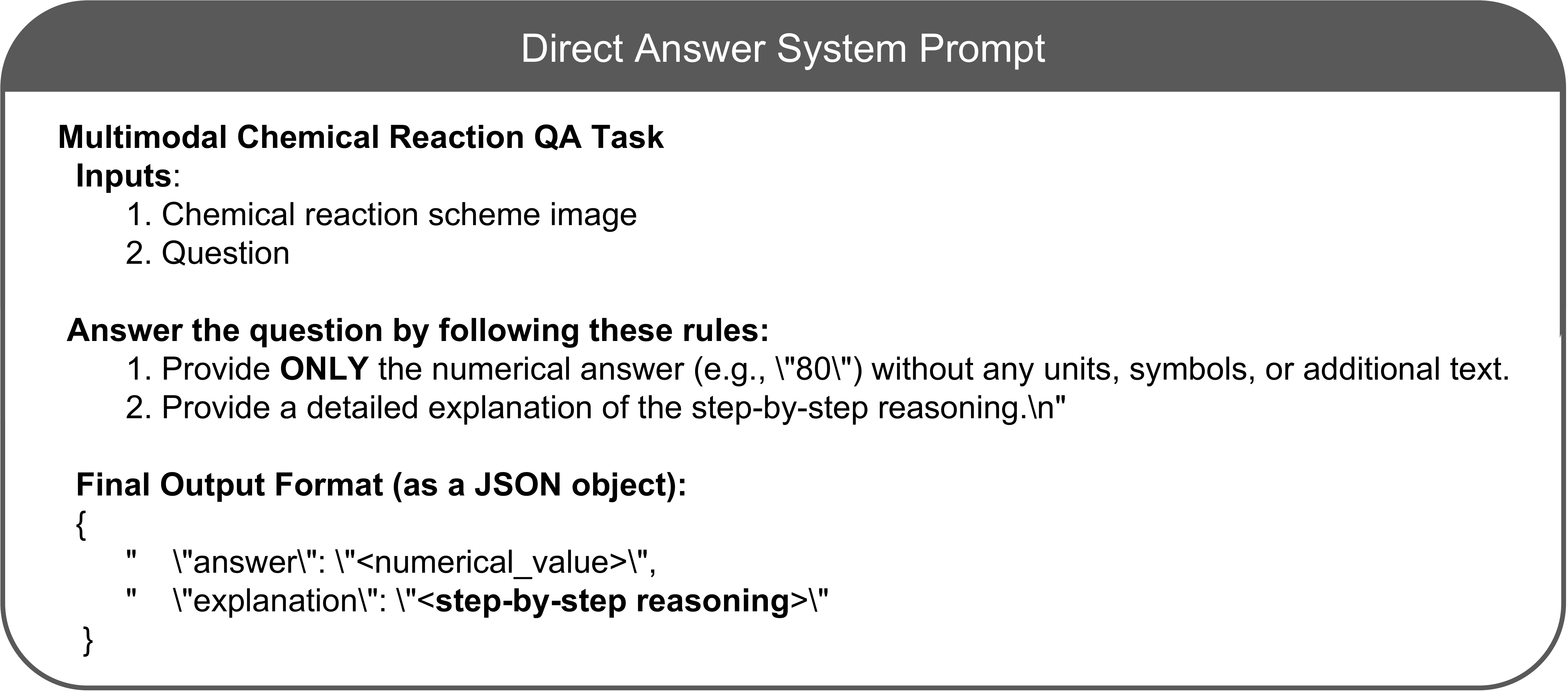}
    \vspace{-5pt}
        \caption{\textbf{The direct answer prompt in our \ours.}}
    \label{fig:dap}
\end{figure*}

In our main experiment, we directly pose questions to the model and employ prompt engineering and template matching to extract answers. As illustrated in Fig.~\ref{fig:dap}, prompts guide the model in generating responses in both full and short answer formats, where the short answer format adheres to the requirements specified in Tab.~\ref{tab:prompt} for different question types. After generation, the short answer is extracted for comparison with the ground truth, while the full answer is used to analyze the reasoning process of MLLMs.

\begin{table}[h]
\caption{The prompt for different QA types in answer generation.}
\label{tab:prompt}
\centering
\small
\begin{tabularx}{\linewidth}{l|X}
\toprule
\textbf{Answer type} & \multicolumn{1}{c}{\textbf{Prompt}} \\
\midrule
multiple choice & Just provide the corresponding choice option, such as `A', `B', `C', or `D'. \\
\midrule
number & Provide ONLY the numerical answer without any units, symbols, or additional text.\\
\midrule
text & Just give the subscript without any other text.\\
\bottomrule
\end{tabularx}
\end{table}

\subsection{Fine-Grained Subtype Statistics}
\label{app:subtype_stats}

Fig.~\ref{fig:stat} reports the proportional composition of the
fine-grained subtypes. Because rounded percentages may obscure the
actual sample sizes of rare categories, we additionally report absolute
counts alongside exact percentages in Tab.~\ref{tab:subtype_stats}.

\begin{table*}[t]
\centering
\small
\setlength{\tabcolsep}{5pt}
\begin{tabular}{llrr}
\toprule
\textbf{Task Dimension} & \textbf{Subtype} & \textbf{Count} & \textbf{\%} \\
\midrule
\multirow{5}{*}{\shortstack[l]{Spatial Element Localization}}
 & Reactant            & 182 & 11.25 \\
 & Product             & 180 & 11.12 \\
 & Reaction Steps      & 416 & 25.71 \\
 & Reaction Conditions &  57 &  3.52 \\
\cmidrule(l){2-4}
 & \textit{Total}      & \textbf{835} & \textbf{51.61} \\
\midrule
\multirow{5}{*}{\shortstack[l]{Topological Information Extraction}}
 & Temperature & 110 & 6.80 \\
 & Time        &  62 & 3.83 \\
 & Yield       &  84 & 5.19 \\
 & Subscript   & 158 & 9.77 \\
\cmidrule(l){2-4}
 & \textit{Total} & \textbf{414} & \textbf{25.59} \\
\midrule
\multirow{4}{*}{\shortstack[l]{Pathway Connectivity Tracing}}
 & Forward Reasoning & 69 & 4.26 \\
 & Reverse Reasoning & 78 & 4.82 \\
 & Path Tracing      & 20 & 1.24 \\
\cmidrule(l){2-4}
 & \textit{Total} & \textbf{167} & \textbf{10.32} \\
\midrule
\multirow{3}{*}{\shortstack[l]{Structural Topology Reasoning}}
 & Topology Classification & 200 & 12.36 \\
 & Cyclicity Detection     &   2 &  0.12 \\
\cmidrule(l){2-4}
 & \textit{Total} & \textbf{202} & \textbf{12.48} \\
\midrule
\textbf{Overall} & & \textbf{1{,}618} & \textbf{100.00} \\
\bottomrule
\end{tabular}
\caption{Fine-grained subtype statistics of ChemReaction. Percentages
are computed over all 1{,}618 QA pairs.}
\label{tab:subtype_stats}
\end{table*}

\noindent\textbf{Imbalance among fine-grained subtypes.}
These statistics make the imbalance across subtypes explicit. Path
Tracing contains 20 examples, while Cyclicity Detection contains only 2.
We note that the Cyclicity Detection subset is too small to support a
statistically reliable standalone evaluation or any broad claim about
models' cycle-detection capabilities; these examples were included to
provide coverage of cyclic reaction structures rather than to establish
an independent quantitative category. Path Tracing, although larger, is
still substantially smaller than Forward and Reverse Reasoning, so we
treat its results as exploratory and refrain from using small
subtype-level differences to support model rankings or general
conclusions.

All quantitative findings reported in this paper are computed at the
level of the four broader task dimensions, which contain 835, 414, 167,
and 202 examples respectively. The subtype-level statistics are intended
to clarify dataset composition and illustrate the coverage of different
operations, not to imply that every subtype supports equally strong
statistical conclusions.

\subsection{Prompt of Chain-of-Thought}
\begin{figure*}[!ht] 
    \centering    
    \includegraphics[width=0.88\linewidth]{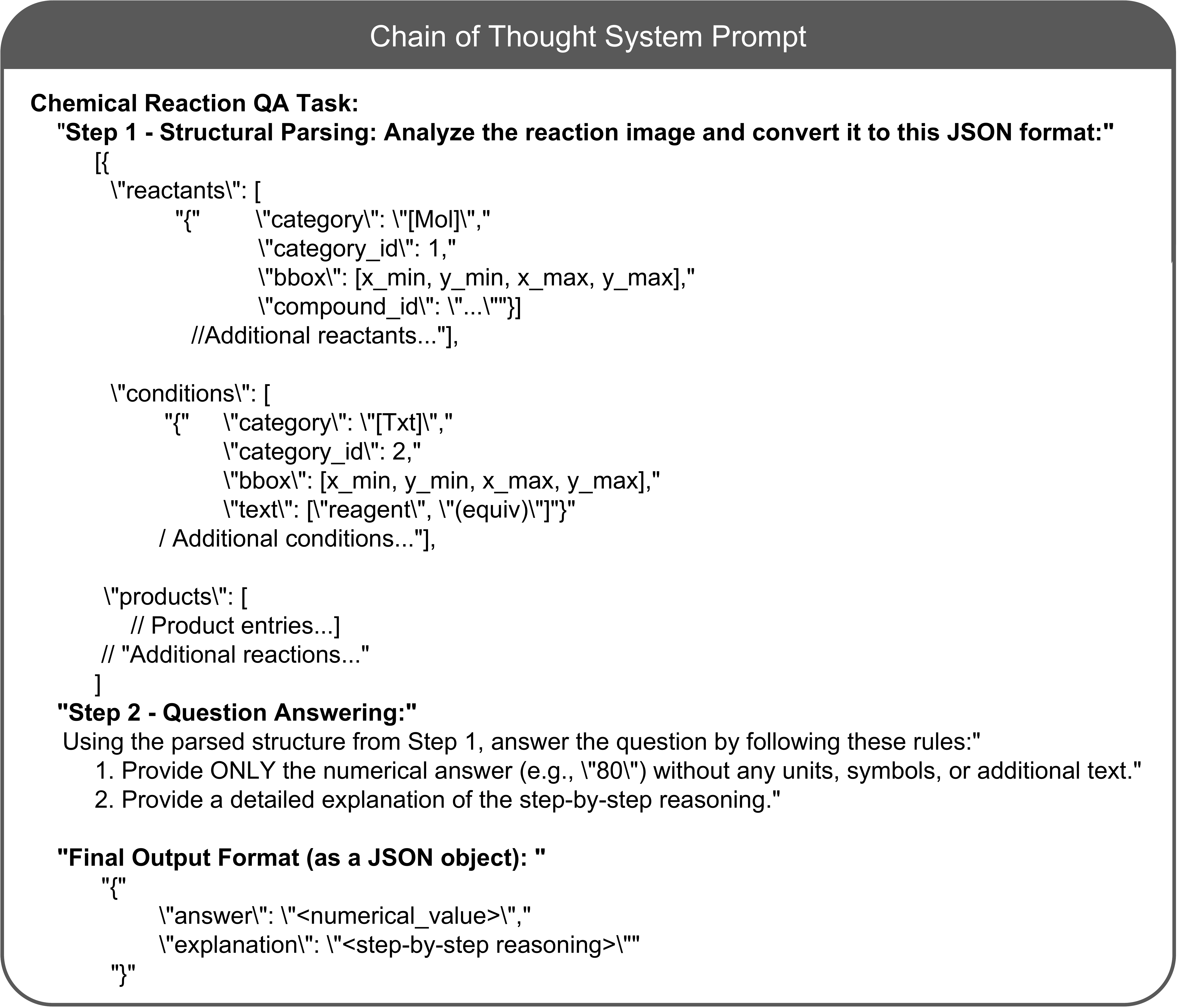}
    \vspace{-5pt}
        \caption{\textbf{The prompt of Chain-of-Thought in our \ours.}}
    \label{fig:cotp}
\end{figure*}

The prompt of the CoT prompting strategy is illustrated in Fig.~\ref{fig:cotp}, where the model is first instructed to convert the reaction diagram into a predefined JSON format, followed by the question being posed.

\subsection{Prompt of External Knowledge}
\label{app:ek}
\begin{figure*}[!ht] 
    \centering
    \includegraphics[width=0.9\linewidth]{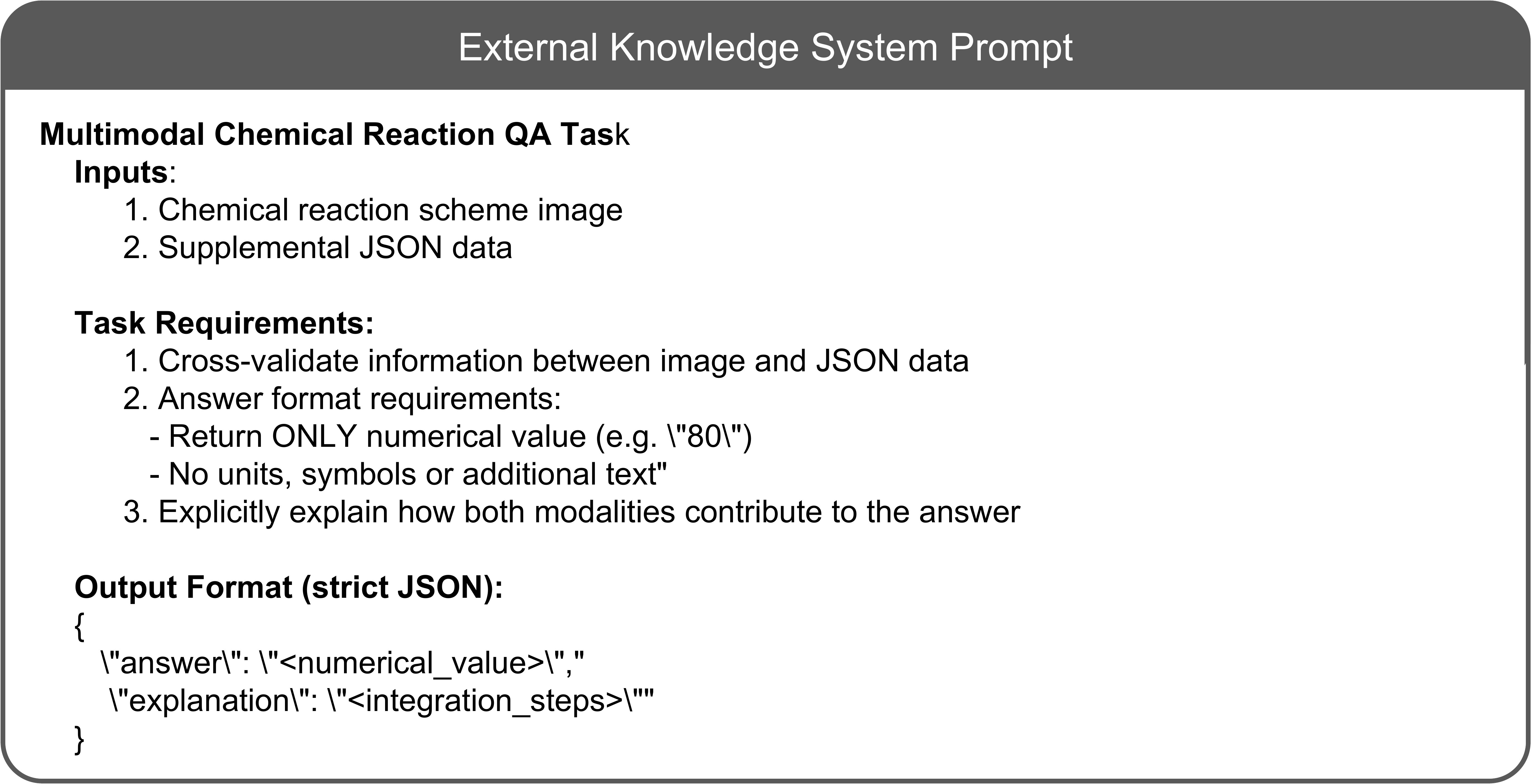}
    \vspace{-5pt}
        \caption{\textbf{The prompt of external knowledge in our \ours.}}
    \label{fig:ekp}
\end{figure*}

\begin{figure*}[h] 
    \centering    
    \includegraphics[width=0.7\linewidth]{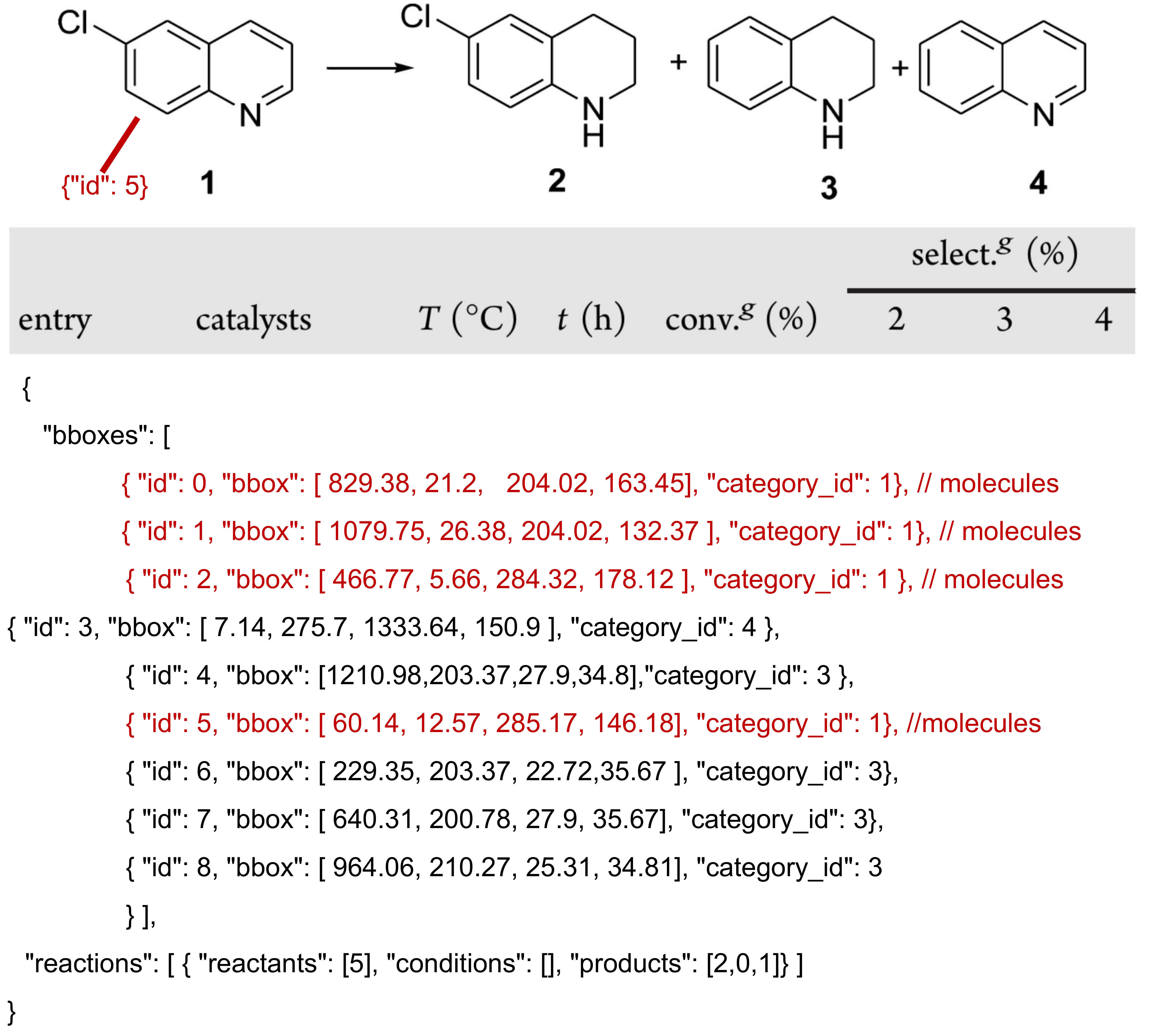}
    \vspace{-5pt}
        \caption{\textbf{External Knowledge Example in our \ours.} This JSON represents molecular structures and reaction relationships extracted from the image. The \texttt{bboxes} section defines detected elements with their respective IDs, bounding box coordinates, and the \texttt{category\_id}, where \texttt{1} indicates molecules, \texttt{2} denotes text, \texttt{3} corresponds to compound identifiers (numerical labels without molecular structures), and \texttt{4} represents auxiliary information. The \texttt{reactions} section associates reactants and products by referencing their IDs, illustrating the transformations occurring between molecules. Overall, it systematically organizes visual and chemical data for structured interpretation.
}
    \label{fig:sql}
\end{figure*}

Here, we present the specific prompt and external knowledge format used
in the main text. The prompt itself is shown in Fig.~\ref{fig:ekp}, and
Fig.~\ref{fig:sql} gives a concrete example of the accompanying JSON.

\begin{figure}[!t]
    \centering  
    \includegraphics[width=0.8\linewidth]{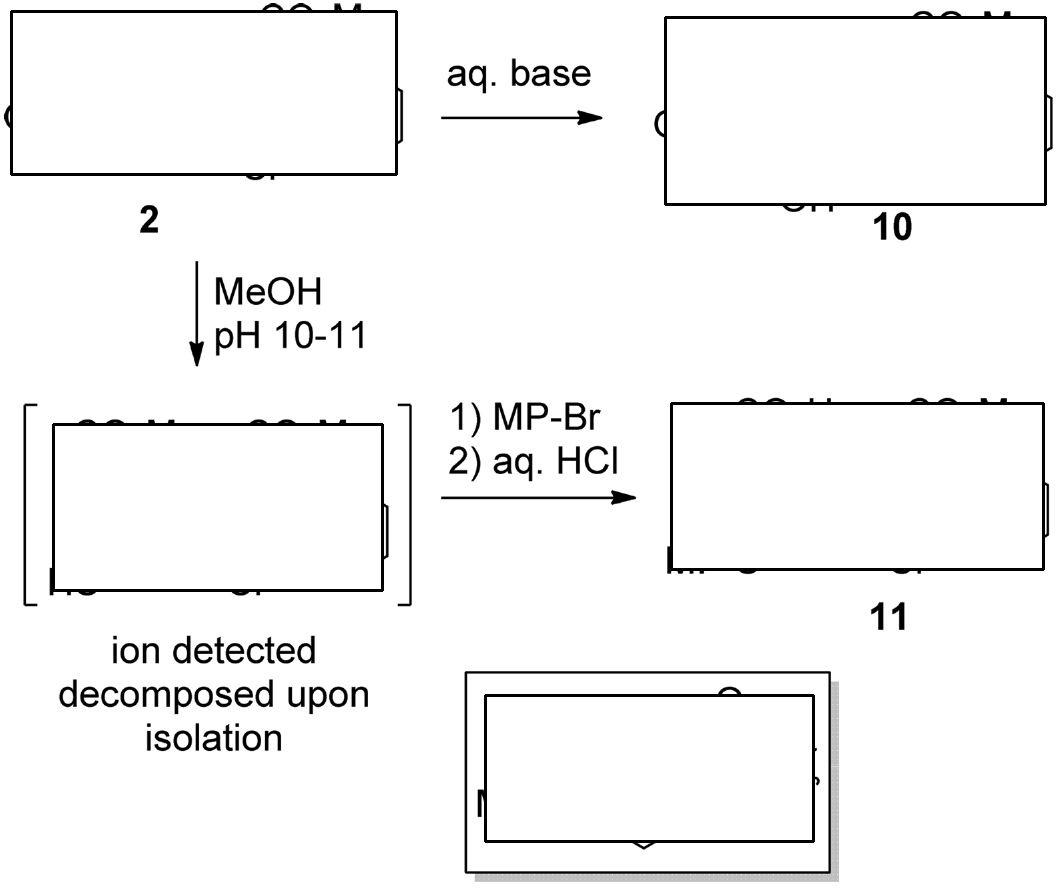}
        \caption{\textbf{Illustration of the visual context ablation.} Detailed molecule images are masked and replaced with rectangle placeholders to isolate the topological structure.}
    \label{fig:ablation}
\end{figure}

\subsection{Illustration of Molecular Structure Masking}
\label{app:masking}
To provide a clear visual example of the visual context ablation described in the main paper, Fig.~\ref{fig:ablation} illustrates how the chemical diagrams are specifically modified for this experiment. By replacing detailed molecular structures with black rectangular placeholders, we force the models to rely solely on the remaining topological elements (e.g., arrows and text) without the aid of familiar chemical visual cues.

\section{Dataset Details}
\label{app:dataset_details}




\subsection{Annotation Protocol and Annotator Qualifications}
We established a rigorous annotation pipeline involving domain experts. All annotators are graduate-level chemists with extensive experience in reading and interpreting organic reaction schemes.
\begin{itemize}
    \item \textbf{Guidelines:} Annotators followed detailed guidelines designed to standardize the extraction of structured information. This included specific instructions on (i) identifying and labeling reactants, products, reagents, and reaction conditions; and (ii) formulating diverse question-answer pairs that target recognition (e.g., entity identification) as well as higher-level reasoning (e.g., mechanistic relations, condition analysis).
    \item \textbf{Training Phase:} Before large-scale annotation, all annotators underwent a training phase using pilot examples. Their initial annotations were reviewed and corrected by a senior expert to ensure alignment with the guidelines.
    \item \textbf{Data Format:} Each reaction diagram is annotated with structured ground truth in JSON format, capturing the semantic roles of all chemical entities within the diagram.
\end{itemize}

\subsection{Quality Control and Inter-Annotator Agreement}
To ensure high data quality, we implemented a multi-stage verification process. Each sample was independently reviewed by three distinct annotators. Disagreements regarding diagram interpretation or answer correctness were resolved through group discussion, with a senior expert making the final decision in ambiguous cases. Through this rigorous review process, the empirical consistency of the final annotations exceeded 95\%.

\section{Examples in \ours}
\label{sec:examples}
\begin{figure*}[h] 
    \centering    
    \includegraphics[width=\linewidth]{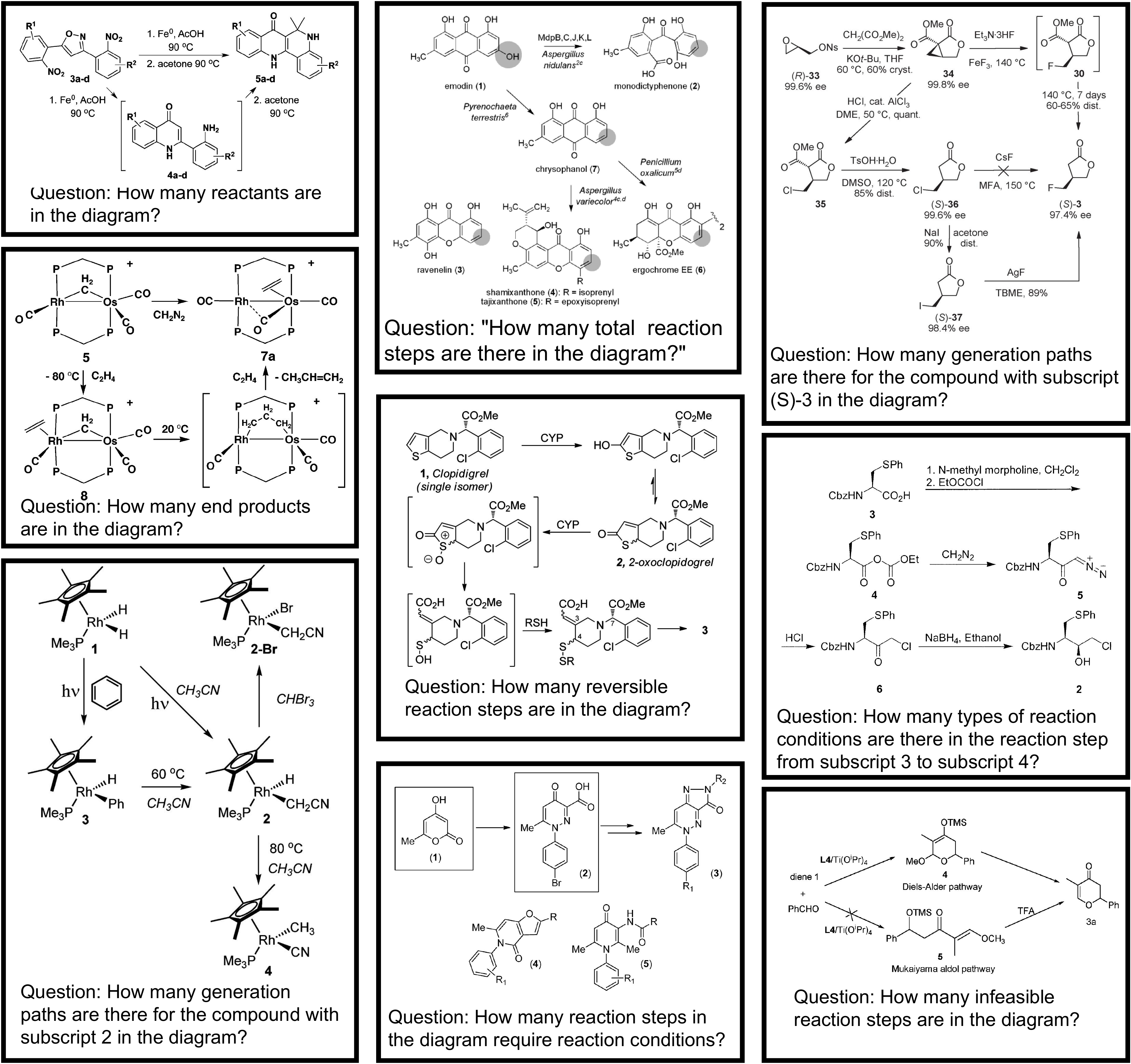}
    \vspace{-5pt}
        \caption{\textbf{Examples of Element Localization Task in our \ours.}}
    \label{fig:class1}
\end{figure*}

\begin{figure*}[h] 
    \centering    
    \includegraphics[width=\linewidth]{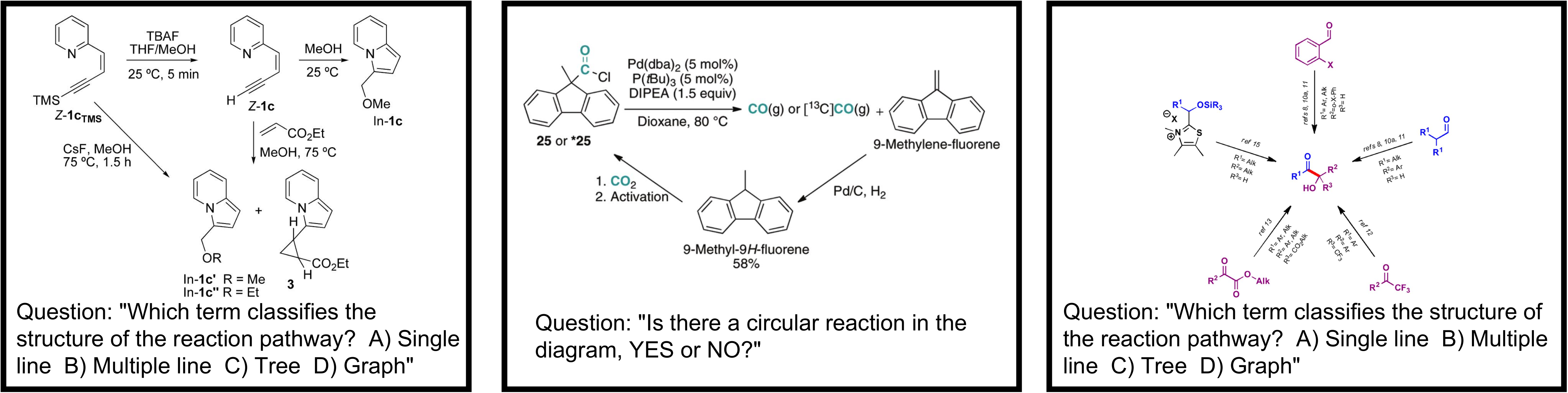}
    \vspace{-5pt}
        \caption{\textbf{Examples of Topology Reasoning Task in our \ours.}}
    \label{fig:class4}
\end{figure*}

\begin{figure*}[h] 
    \centering    
    \includegraphics[width=\linewidth]{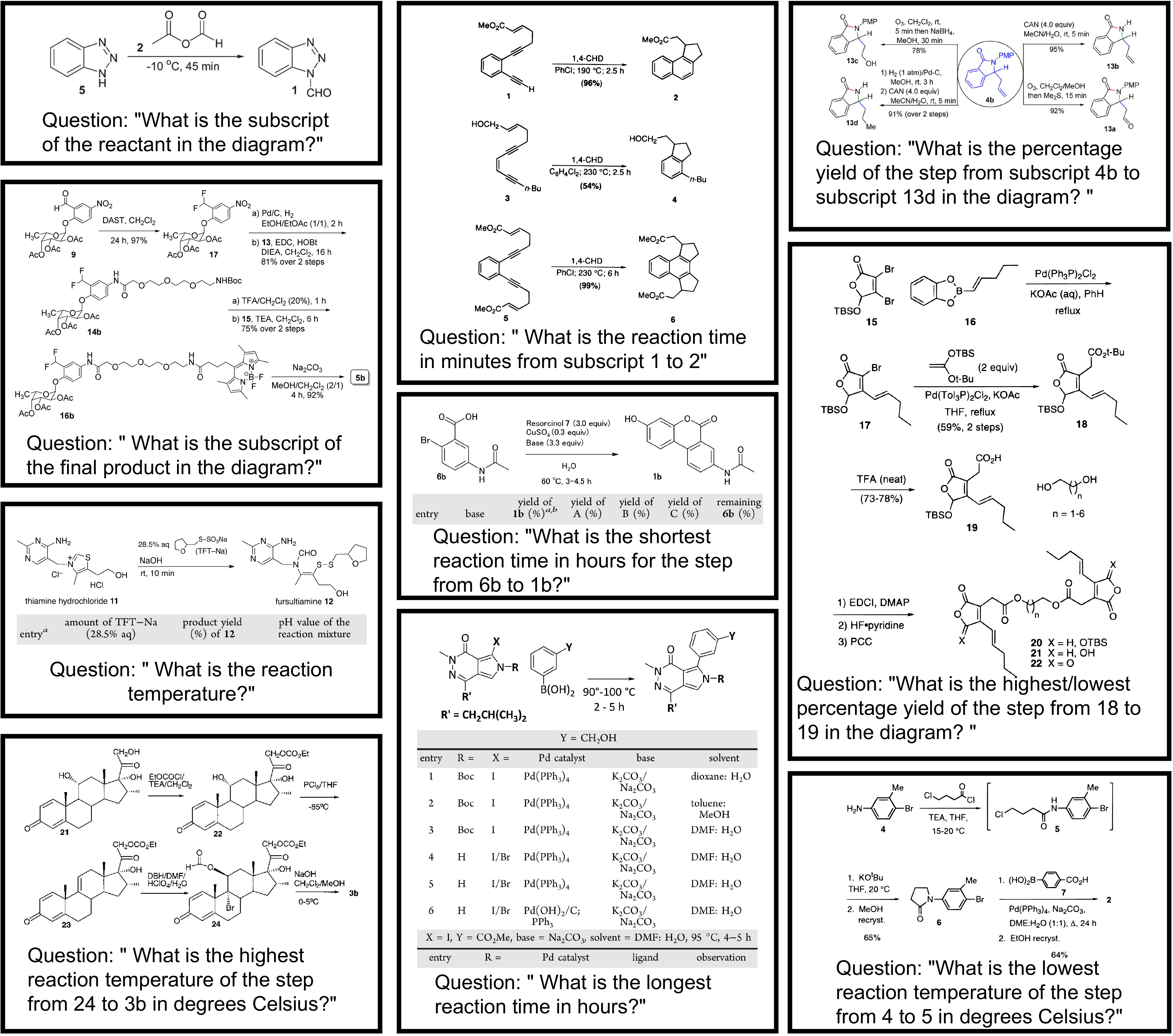}
    \vspace{-5pt}
        \caption{\textbf{Examples of Information Extraction Task in our \ours.}}
    \label{fig:class2}
\end{figure*}

\begin{figure*}[h] 
    \centering    
    \includegraphics[width=\linewidth]{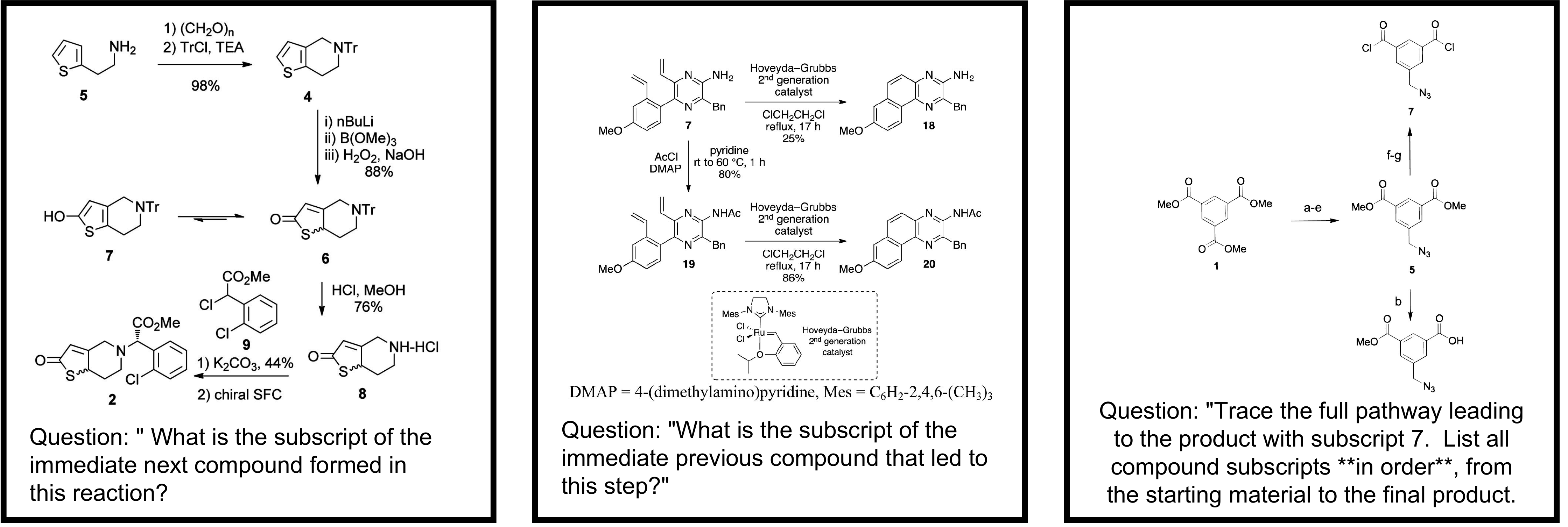}
    \vspace{-5pt}
        \caption{\textbf{Examples of Connectivity Tracing Task in our \ours.}}
    \label{fig:class3}
\end{figure*}

In this section, we present selected samples from different tasks in our dataset, as illustrated in Fig.~\ref{fig:class1}, ~\ref{fig:class4}, ~\ref{fig:class2}, and ~\ref{fig:class3}. These examples highlight the diversity of our designed questions, which encompass a wide range of problem types.

\end{document}